\documentclass[dvipsnames,format=sigconf,anonymous=false,review=false]{acmart} 

\def\vec#1{\mathchoice{\mbox{\boldmath$\displaystyle#1$}}
{\mbox{\boldmath$\textstyle#1$}}
{\mbox{\boldmath$\scriptstyle#1$}}
{\mbox{\boldmath$\scriptscriptstyle#1$}}}
\usepackage{booktabs} % For formal tables
\usepackage{amsmath}
\usepackage{algorithm}
\usepackage[noend]{algpseudocode}
\usepackage[T1]{fontenc}
\usepackage{tabularx}
\usepackage{multirow}
\usepackage{array}
\usepackage{pbox}

\usepackage{color,soul}
\usepackage{nameref}

\usepackage{pgfplots}
\usepackage{graphicx}
\usepackage{subcaption}
\usepackage{tikz}
\usetikzlibrary{intersections, pgfplots.fillbetween}
\usepackage{siunitx}

\usepackage{url}
\usepackage{bm}

\definecolor{darkgreen}{rgb}{0.30, 0.50, 0.0}

\makeatletter
\renewcommand{\ALG@name}{Pseudocode}
\makeatother

\newcolumntype{L}[1]{>{\raggedright\let\newline\\\arraybackslash\hspace{0pt}}m{#1}}
\newcolumntype{C}[1]{>{\centering\let\newline\\\arraybackslash\hspace{0pt}}m{#1}}
\newcolumntype{R}[1]{>{\raggedleft\let\newline\\\arraybackslash\hspace{0pt}}m{#1}}

\newtheorem{thm}{Theorem}

\newcommand{\VIG}{G}
\newcommand{\VIGnl}{G_{NL}}
\newcommand{\VIGnm}{G_{NM}}

\newcommand{\eVIG}{eG}

\newcommand{\eVIGnm}{eG_{NM}}

\graphicspath{{images/}}

\copyrightyear{2026}
\acmYear{2026}
\setcopyright{cc}
\setcctype{by-nc-nd}
\acmConference[GECCO '26]{Genetic and Evolutionary Computation Conference}{July 13--17, 2026}{San Jose, Costa Rica}
\acmBooktitle{Genetic and Evolutionary Computation Conference (GECCO '26), July 13--17, 2026, San Jose, Costa Rica}
\acmDOI{10.1145/3795095.3805171}
\acmISBN{979-8-4007-2487-9/2026/07}

\hypersetup{draft}
\begin{document}
	%\title[No-cost missing linkage detection and local search supported by the limited non-monotonical surrogate constructed using recursive empirical linkage learning techniques]{No-cost missing linkage detection and local search supported by the limited non-monotonical surrogate constructed using recursive empirical learning techniques}
    %\title[aaa]{On constructing Limited Perfect Monotonical Surrogates via low-cost recursive empirical linkage discovery with guaranteed output}
    \title[Limited Monotonical Perfect Surrogates]{Limited Perfect Monotonical Surrogates constructed using low-cost recursive linkage discovery with guaranteed output}

 	\author{Michal W. Przewozniczek}
 	\affiliation{
            %\institution{Dep. of Systems and Comp. Networks}
 		\institution{Wroclaw Univ. of Science and Techn.}
 		\city{Wroclaw}
 		\country{Poland} 
 	}
 	\email{michal.przewozniczek@pwr.edu.pl}

        \author{Francisco Chicano}
        \affiliation{
        %\institution{Dep. of Lang. and Comp. Sciences}
          \institution{ITIS Software, University of M\'alaga}
         \city{M\'alaga} 
         \country{Spain} 
         \postcode{29071}
        }
        \email{chicano@uma.es}

        \author{Renato Tin\'os}
        \affiliation{
        %\institution{Dep. of Computing and Mathematics}
        \institution{University of S\~ao Paulo}
        \city{Ribeir\~ao Preto} 
        %\STATE{S\~ao Paulo}
        \country{Brazil}
        \postcode{14040901}
        }
        \email{rtinos@ffclrp.usp.br}

        \author{Marcin M. Komarnicki}
 	\affiliation{
 	%\institution{Dep. of Systems and Comp. Networks}
 		\institution{Wroclaw Univ. of Science and Techn.}
 		\city{Wroclaw}
 		\country{Poland} 
 	}
 	\email{marcin.komarnicki@pwr.edu.pl}

	\renewcommand{\shortauthors}{Michal W. Przewozniczek et al.}

	\begin{abstract}
        Surrogates provide a cheap solution evaluation and offer significant leverage for optimizing computationally expensive problems. Usually, surrogates only approximate the original function. Recently, the perfect linear surrogates were proposed that ideally represent the original function. These surrogates do not mimic the original function. In fact, they are another (correct) representation of it and enable a wide range of possibilities, e.g., discovering the optimized function for problems where the direct transformation of the encoded solution into its evaluation is not available. However, many real-world problems can not be represented by linear models, making the aforementioned surrogates inapplicable. Therefore, we propose the Limited Monotonical Perfect Surrogate (LyMPuS), which overcomes this difficulty and enables the comparison of two solutions that differ by a single variable. Our proposition is suitable for limiting the cost of expensive local search procedures. The proposed surrogate is parameterless and can be trained on the fly without any separate surrogate-building step. It uses only the necessary fitness evaluations, and the already-paid costs are not wasted when the model is updated. Finally, it offers low-cost missing-linkage detection and low-cost linkage discovery, guaranteed to find a missing dependency in no more than $2\lceil\log_2(n)\rceil$ steps.
        \end{abstract}
	
	%
	% The code below should be generated by the tool at
	% http://dl.acm.org/ccs.cfm
	% Please copy and paste the code instead of the example below. 
	%
    

	

	
	\maketitle
	
	%\textcolor{red}{\textbf{The supplementary material for this document is available at}}\\
%\textcolor{red}{\url{http://www.przewozniczek.eu/download/weightWalshSUPPL.pdf}}

\section{Introduction}
\label{sec:intro}

The evaluation of many real-world problems is expensive and surrogates can significantly leverage the effectiveness and efficiency of the optimizers applied to solve them \cite{verelSparseWalshModels}. Typically, surrogates aim at approximating the optimized function \cite{surrStandard}. Recently, perfect surrogates were proposed for the binary \cite{walshSurBin,ellGomea} and permutation-based problems \cite{walshSurPerm}. Obviously, perfect surrogates are equivalent to evaluating the original function and have a significant advantage over the typical surrogates that only approximate solution evaluation. However, their capabilities are significantly broader. Many real-world problems require indirect solution encoding; e.g., a simulation must be executed to evaluate the solution, or the encoding serves as input to the solution-building algorithm \cite{automated_timetabling,multi-echelon}. In such cases, we know the exact function that evaluates the final solution, e.g., a production plan. However, we usually do not know the function that is optimized, i.e., one that directly transforms the encoded solution into its evaluation. Perfect surrogates allow us to overcome this difficulty and discover an optimized function \cite{verelSparseWalshModels}.\par

Perfect surrogates require the non-linearity check to identify the variable dependencies needed to build their internal function model. For many problems, the non-linearity check is oversensitive \cite{GoldMonotonicity}, making the construction of a perfect surrogate intractable. Recent research shows that the non-linearity check is inappropriate for decomposing many real-world problems, which require using the non-monotonicity check \cite{wVIG}. Finally, variable dependency discovery, the first step of the perfect surrogate building procedure, often resembles a blind-luck search and must be performed before surrogate training, making the whole process cumbersome \cite{ellGomea}.\par

To fill the aforementioned gaps, we propose the Limited Monotonical Perfect Surrogate (LyMPuS) that (if trained on the complete set of dependencies obtained using the non-monotonicity check) is guaranteed to compare the quality of two solutions differing by one variable. Such a functionality is suitable to leverage expensive local search procedures. The proposed surrogate is flexible; it is trained on the fly, using fitness evaluations only when necessary, and avoids wasting any of the already-paid costs. It supports low-cost missing linkage detection and the low-cost linkage discovery that is guaranteed to find one missing dependency in no more than $2\lceil\log_2(n)\rceil$ steps, where $n$ is the problem size.

\section{Background}
\label{sec:relWork}

In many combinatorial problems some subsets of variables may have a joint non-linear or non-monotonical influence on function value and, therefore, should be processed jointly \cite{whitleyNext,wVIG}. Here, we consider the maximization of $f(\vec{x})$, where $\vec{x}=[x_1,x_2,\ldots,x_{n}]$ is a binary solution vector of size \textit{n}. However, the mechanisms proposed in this work apply to other search spaces as well.
\example \label{ex:depBasic}
Consider $f(x_1,...,x_8) = f_1(x_1,...,x_4) + f_2(x_5,...,x_8)$, where each $f_i(x_{4i-3},...,x_{4i}) = bim_4(u(x_{4i-3},...,x_{4i}))$, where $bim_4$ is bimodal deceptive function or order $k$ \cite{decBimodalOld}:
\begin{equation}
    \label{eq:decBimod}
    \mathit{bim}_k\bigl(u(\vec{x})\bigr) = 
    \begin{cases}
        k / 2 - |u - k/2| - 1 & ,u \neq k \land u \neq 0\\
        k / 2 & ,u = k \lor u = 0\\
    \end{cases}
\end{equation}
where $u$ is the sum of gene values (so-called \textit{unitation}).\\
Intuitively variables in groups $\{x_1,...,x_4\}$ and $\{x_5,...,x_8\}$ depend on each other and each group can be processed separately to find the optimal solution.
\endexample
We can identify variable dependencies by checking the results of variable perturbation. We define $\Delta_g(\vec{x}) = f(\vec{x^g}) - f(\vec{x})$ and $\Delta_{g,h}(\vec{x}) = f(\vec{x^{g,h}}) - f(\vec{x})$, where $\vec{x^{g}}$ equals $\vec{x}$ with $x_g$ flipped, and $\vec{x^{g,h}}$ equals $\vec{x}$ with $x_g$ and $x_h$ flipped. If, for a given $\vec{x}$, flipping $x_g$ and $x_h$ causes fitness change of $\Delta_g$ and $\Delta_h$, respectively, then flipping them both should modify fitness by $\Delta_{g,h}$. Otherwise, $x_g$ and $x_h$ are \textit{non-linearly} dependent. In the \textit{non-linearity check} \cite{linc}, $x_g$ and $x_h$ are dependent if: $f(\vec{x}) + f(\vec{x}^{g,h}) \neq f(\vec{x}^g) + f(\vec{x}^h)$. In Example \ref{ex:depBasic}, in two separate groups, $\{x_1,...,x_4\}$ and $\{x_5,...,x_8\}$, variables are non-linearly dependent on each other.\par

Any problem can be represented in the \textit{additive form}: $f(\vec{x})=\sum_{s=1}^{S} f_s(\vec{x}_{I_s})$, where the argument set $\vec{x}_{I_s}$ of each subfunction $f_s$ is a subset of $\vec{x}$, and $\vec{x}_{I_s}$ do not have to be disjoint. The non-linearity check is convenient for decomposing the \textit{k}-bounded problems that can be represented in such an additive form that each subfunction takes no more than $k$ arguments.\par

The non-linearity check is oversensitive for many real-world problems \cite{GoldMonotonicity,wVIG}. Using $f(...)$ from Example \ref{ex:depBasic}, we can define $f'(x_1,...,x_8) = \bigl(f(x_1,...,x_8)\bigr)^2$. For any two solutions $\vec{x_a}$ and $\vec{x_b}$, it is true that if $f(\vec{x_a}) R f(\vec{x_b})$, then $f'(\vec{x_a}) R f'(\vec{x_b})$ where $R \in \{>,<,=\}$. Thus, $f$ and $f'$ have the same local and global optima and their attraction basins. However, the non-linearity check will recognize all variables in $f'$ as dependent on each other, which is useless information. To this end, the non-monotonicity check was proposed \cite{GoldMonotonicity}. We present its slightly more sensitive version \cite{2dled}, i.e., $x_g$ and $x_h$ are non-monotonically dependent if at least one clause holds:\\
\footnotesize
    \textbf{C1.} $f(\vec{x}) < f(\vec{x}^g)  \land f(\vec{x}^h) \geq f(\vec{x}^{g,h})$   \textbf{C4.} $f(\vec{x}) < f(\vec{x}^h)  \land f(\vec{x}^g) \geq f(\vec{x}^{g,h})$\\
    \textbf{C2.} $f(\vec{x}) = f(\vec{x}^g)  \land f(\vec{x}^h) \neq f(\vec{x}^{g,h})$    \textbf{C5.} $f(\vec{x}) = f(\vec{x}^h)  \land f(\vec{x}^g) \neq f(\vec{x}^{g,h})$\\
    \textbf{C3.} $f(\vec{x}) > f(\vec{x}^g)  \land f(\vec{x}^h) \leq f(\vec{x}^{g,h})$   \textbf{C6.} $f(\vec{x}) > f(\vec{x}^h)  \land f(\vec{x}^g) \leq f(\vec{x}^{g,h})$\\
\normalsize
Consider clause C1 -- flipping $x_g$ improves fitness. However, if $x_h$ is flipped, flipping $x_g$ does not improve fitness anymore. Thus, $x_g$ depends on $x_h$. Some recent studies show that the non-monotonicy check can discover non-symmetrical relations (i.e., $x_g$ depends on $x_h$ but $x_h$ does not have to depend on $x_g$) and that telling them from symmetrical relations can cause a significant performance improvement while optimizing some real-world problems \cite{2dled}. Here, we focus on symmetrical dependencies, but all proposed mechanisms also apply to discovering non-symmetrical dependencies.\par

In complex real-world problems, the subfunction argument sets frequently overlap, i.e., share variables \cite{conflictingOverlapsAreRealWorld,wVIG}. Such overlapping problems are hard to optimize because, in the general case, subfunctions can not be optimized separately. Gray-box optimization, where the variable dependencies are known \textit{a priori}, proposes effective operators to handle overlapping problems \cite{whitleyNext}. A variable dependency network is frequently represented by the Variable Interaction Graph (VIG, $\VIG$). $\VIG$ contains a complete set of true dependencies of a given type, e.g., non-linear ($\VIGnl$) or non-monotonical ($\VIGnm$). For any $\vec{x}$ and $x_g$, using $\VIG$ (of any type), one can construct a perturbation mask $m$ containing (all or a subset of) variables dependent on $x_g$. After perturbing, $\vec{x}$ is the subject of local optimization yielding $\vec{x'}$. If $f(\vec{x'}) \geq f(\vec{x})$, then $\vec{x'}$ replaces $\vec{x}$ or is rejected otherwise. Such simple, yet dependency-driven, operators were shown to be effective in solving many complex overlapping problems \cite{ilsDLED}.\par

Partition Crossover (PX) is a gray-box mixing operator \cite{pxForBinary}. For two mixed individuals, say $\vec{x_a}$ and $\vec{x_b}$, it clusters all variables differing in $\vec{x_a}$ and $\vec{x_b}$ with respect to $\VIG$. PX was shown to produce mixing masks that preserve subfunction argument sets even if they overlap \cite{dgga}. For $\VIGnl$, any mask yielded by PX is proven to yield offspring such that $f(\vec{x_a}) + f(\vec{x_b}) = f(\vec{x_{oa}}) + f(\vec{x_{ob}})$ \cite{pxLatticesHyper}. For $\VIGnm$, PX is proven to yield offspring such that:
\begin{itemize}
    \item \textbf{if} $f(\vec{x_a}) R_{ine} f(\vec{x_{oa}})$ \textbf{then} $f(\vec{x_b}) \neg R_{ine} f(\vec{x_{ob}})$, where $R_{ine} = \{>,<\}$,  $\vec{x_{oa}} = \vec{x_a} \xleftarrow[mPX]{} \vec{x_b}$ ($\vec{x_{oa}}$ is a copy of $\vec{x_a}$ with gene values marked by mask $mPX$ copied from $\vec{x_b}$), and $mPX$ is any PX mask yielded for $\vec{x_a}$ and $\vec{x_b}$.
    \item \textbf{if} $f(\vec{x_a}) = f(\vec{x_{oa}})$ \textbf{then} $f(\vec{x_b}) = f(\vec{x_{ob}})$.\par
\end{itemize}

In black-box optimization, variable dependencies are unknown. Among all, we can discover them using non-linearity or non-mono-tonicity checks, store them in the empirical VIG ($\eVIG$), and use them in the gray-box manner. All dependencies discovered using such checks are true. Thus, $\eVIG \subseteq \VIG$. PX can be used for individual mixing and missing linkage detection, i.e., when the offspring does not fulfill the aforementioned features, then we know the PX mask is incomplete. Thus, there must exist at least one variable from the PX mask and one from the rest of the genotype that are dependent \cite{dgga}. See Section S-I (supplementary material) for PX examples.\par

%Non-monotonicity check was frequently considered as poorly sensitive \textcolor{red}{[BIB]}. The recent propositions concerning various search spaces have significantly limited its costs \cite{ilsDLED,FIHCwLL,irrg,dgga}. Nevertheless, none of these linkage learning techniques guarantees discovering the missing dependency if there are any.\par

Any pseudo-Boolean function can be represented as $f(\vec{x}) = \sum_{i=0}^{2^n-1} w_i \varphi_i(\vec{x})$ where $w_i \in \mathbb{R}$ is the $i$th Walsh coefficient, $\varphi_i(\mathbf{x}) =(-1)^{\mathbf{i}^\mathrm{T}\mathbf{x}}$ generates a sign, and $\mathbf{i} \in \{0,1\}^n$ is the binary representation of index $i$. The Walsh-based model of $f(\vec{x})$ can be constructed in black-box optimization \cite{ellGomea,walshSurBin2,walshSurPerm}. To this end, one have to discover non-linear dependencies and use the Walsh-based model construction algorithm \cite{heckendorn2002}. If all non-linear dependencies were discovered, then the Walsh-based model would perfectly represent the optimized function, allowing for evaluating it using the obtained \textit{Perfect Linear Surrogate} rather than the original function. Such surrogates enable much more than just free fitness evaluations. For instance, when $\vec{x}$ defines a entry information to some solution-building algorithm, e.g., in feature selection problems \cite{wVIG}, then we frequently do not know the function that directly transforms $\vec{x}$ into its evaluation (in feature selection problems we usually know a function that evaluates an encoded classifier but not a function that evaluates the encoding). Perfect Surrogates allow to break this limitation.\par

Both non-linearity and non-monotonicity checks are applicable for continuous problems as well~\cite{irrg,mdg,rdg3}. Flipping a variable value is then equivalent to setting a different value within its domain. Problems are mainly first decomposed, and each part is optimized separately afterward. In this approach, parts are disjoint, which may not be suitable for overlapping problems~\cite{occ}. To reduce the decomposition cost, recursive decomposition strategies are used~\cite{irrg,mdg,rdg3}. Most of them group all variables together when overlapping problems are considered~\cite{rdg3}. Moreover, searching for dependencies for a single variable starts with the interaction check with all variables that do not yet belong to any group.\par

Here, we consider optimizers using Statistical Linkage Learning (SLL) \cite{ltgaOriginal,P3Original,dsmga2} supported by the recently proposed First Improvement Hillclimber with Linkage Learning (FIHCwLL) \cite{FIHCwLL}. Since these optimizers are only used to present the quality of our propositions, they are described in Section S-II (supplementary material).

\section{Limited Monotonic Perfect Surrogate}
\label{sec:LNMSurr}

\subsection{Motivations and ideas}
\label{sec:LiPMos:ideas}
%\textcolor{red}{define: context of a variable (show its name in the recent Darrell's work)}\\
%\textcolor{red}{consequence only the context influences the result of a given variable flip}\\
%\textcolor{red}{present LiPMoS}\\

Assume that $\VIG$ is known. We define $C_{\VIG}^g$, the \textit{close context} of $x_g$ as the set of all variables adjacent to $x_g$ in $\VIG$. 
%i.e., $C_{\VIG}^g = \{x_{i_1},...,x_{i_m}\}$ such that $\forall {x_{i_k}} \in C_{\VIG}^g G(x_g, x_{i_k}) = 1$ and $\nexists x_i \notin C_{\VIG}^g G(x_g, x_i) = 1$. 
By $C_{\VIG}^g(\vec{x})$, we define a hyperplane \cite{pxLatticesHyper} 
%(based on $C_{\VIG}^g$ and $\vec{x}$) 
such that the values of $C_{\VIG}^g$ are taken from $\vec{x}$. The set of variables marked by a hyperplane $\phi$ we define as $def(\phi)$. For instance, for $\vec{x} = [111000]$ and $C_{\VIG}^2 = \{x_1,x_3, x_6\}$, $C_{\VIG}^2(\vec{x}) = [\text{1*1**0}]$, where * indicates the \textit{don't care} value. Finally, $def([\text{1*1**0}]]) = C_{\VIG}^2$.

\par

\begin{thm}
    \label{thm:flipping}
        Consider any $\vec{x_a}$, $\vec{x_b}$ such that $x_g$ equals in $\vec{x_a}$, $\vec{x_b}$,  $C_{\VIG}^g(\vec{x_a}) = C_{\VIG}^g(\vec{x_b})$, and $R_1, R_2\in\{<,=,>\}$ such that $f(\vec{x_a}) R_1 f(\vec{x^g_a})$, $f(\vec{x_b}) R_2 f(\vec{x^g_b})$. If $\VIG = \VIGnl$ or $\VIG = \VIGnm$, then $R_1=R_2$.
    \begin{proof}
        If $\VIG = \VIGnl$, then $\Delta_g(\vec{x_a}) \neq \Delta_g(\vec{x_b})$ if and only if at least one gene differing in $\vec{x_a}$ and $\vec{x_b}$ is non-linearly dependent on $x_g$. Since $C_{\VIGnl}^g(\vec{x_a}) = C_{\VIGnl}^g(\vec{x_b})$, then all genes non-linearly dependent on $x_g$ are equal in $\vec{x_a}$ and $\vec{x_b}$. Thus, $\Delta_g(\vec{x_a}) = \Delta_g(\vec{x_b})$ and $R_1=R_2$.\par
        If $\VIG = \VIGnm$, then $R_1 \neq R_2$ if and only if one of the C1-C6 clauses holds for $x_g$ and at least one other gene that differs in $\vec{x_a}$ and $\vec{x_b}$. Since, $C_{\VIGnm}^g(\vec{x_a}) = C_{\VIGnm}^g(\vec{x_b})$, then all genes 
        %non-monotonically dependent on $\vec{x_g}$
        for which at least one of C1-C6 clasuses is satisfied for $\vec{x_g}$ are equal in $\vec{x_a}$ and $\vec{x_b}$. Thus, $R_1=R_2$.
    \end{proof}
\end{thm}

 \begin{table}  
    \caption{$\VIGnm$ and LyMPuS for $f_{e1}$.} 
    \label{tab:Ex:fe1}
    \scriptsize
    \begin{subtable}{.18\textwidth}
    \centering 
       \begin{tabular}{l|cccc}
              & $x_1$  & $x_2$  & $x_3$ & $x_4$  \\
              \hline
            $x_1$ & X & 1 & 0 & 0 \\
            $x_2$ & 1 & X & 1 & 0 \\
            $x_3$ & 0 & 1 & X & 1 \\
            $x_4$ & 0 & 0 & 1 & X  
        \end{tabular}
    \caption{$\VIGnm$ for Example \ref{ex:lipmos}}
    \label{tab:Ex:fe1:VIGnm}
    \end{subtable}
    \begin{subtable}{.25\textwidth}
    \centering 
       \begin{tabular}{l|ccc}
            $x_g$ & $C_{\VIGnm}^g$  & $C_{\VIGnm}^g(\vec{x})$  & $b(x_g, C_{\VIG}^g(\vec{x}), \vec{x})$ \\
              \hline
            $x_1$ & $\{x_2\}$ & *0** & $\{1\}$ \\
                  &           & *1** & $\{0\}$ \\
                  \hline
            $x_2$ & $\{x_1,x_3\}$ & 0*0* & $\{1\}$ \\
                  &               & 0*1* & $\{0,1\}$ \\
                  &               & 1*0* & $\{0,1\}$ \\
                  &               & 1*1* & $\{0\}$ \\
                  \hline
            $x_3$ & $\{x_2,x_4\}$ & *0*0 & $\{1\}$ \\
                  &               & *0*1 & $\{0,1\}$ \\
                  &               & *1*0 & $\{0,1\}$ \\
                  &               & *1*1 & $\{0\}$ \\
                  \hline
            $x_4$ & $\{x_3\}$ & **0* & $\{1\}$ \\
                  &           & **1* & $\{0\}$ \\
        \end{tabular}
    \caption{LyMPuS for Example \ref{ex:lipmos}}
    \label{tab:Ex:fe1:LiPMoS}
    \end{subtable}
\end{table}

For any $x_g$ and a hyperplane $\phi$ such that $x_g \notin def(\phi)$ we define
\begin{equation}
    \label{eq:bHyp}
    b(x_g, \phi, \vec{x}) = 
    \begin{cases}
        \{0\} & \text{ \textbf{if} } f(\vec{x}|_{x_g=0, def(\phi)=\phi}) > f(\vec{x}|_{x_g=1, def(\phi)=\phi}) \\
        \{1\} & \text{ \textbf{if} } f(\vec{x}|_{x_g=0, def(\phi)=\phi}) < f(\vec{x}|_{x_g=1, def(\phi)=\phi}) \\
        \{0,1\} & \text{ \textbf{if} } f(\vec{x}|_{x_g=0, def(\phi)=\phi}) = f(\vec{x}|_{x_g=1, def(\phi)=\phi}) \\
    \end{cases}
\end{equation}
where $\vec{x}|_{x_g=v, def(\phi)=\phi}$ is a solution obtained from $\vec{x}$ by setting the value of $x_g$ to $v$ and the value of all $x_i \in def(\phi)$ to $\phi$, e.g., for $\phi = [\text{***111}]$, $[\text{111000}]|_{x_3 = 0, def{\phi} = \phi} = [\text{110111}]$.\par

Note that for any $\vec{x}$, $b(x_g, \cdot, \vec{x})$ (where $\cdot$ is a hyperplane with no defined positions, i.e., $[\text{******}]$) indicates if flipping $x_g$ is an improving, deteriorating or a sliding move for $\vec{x}$.\par

From Theorem \ref{thm:flipping} it follows that for each $x_g$ and a $\phi$ such that $def(\phi) = C_{\VIG}^g$, it is true that $\forall_{\vec{x_a},\vec{x_b}} b(x_g, \phi, \vec{x_a}) = b(x_g, \phi, \vec{x_b})$. We define the Limited Monotonic Perfect Surrogate (LyMPuS) as
\begin{equation}
    \label{eq:LiPMoS}
    LyMPuS(x_g, \VIG, \vec{x}) = b(x_g, C_{\VIG}^g(\vec{x}), \vec{x})
\end{equation}
If $\VIG$ is known, then to construct LyMPuS it is enough to collect and store the outputs of $b(x_g, C_{\VIG}^g(\vec{x}), \vec{x})$ for all $\phi$ such that $def(\phi) = C_{\VIG}^g$ for each $x_g$. Then, we can use LyMPuS for checking if flipping $x_g$ for any solution $\vec{x}$ improves, decreases or leaves its fitness unchanged. If LyMPuS is constructed for $\VIGnm$, it is convenient for problems that can can not be handled by Walsh-based surrogates.\par

\example \label{ex:lipmos}
Consider $f(x_1,...,x_4) = f_1(x_1,x_2) \cdot f_2(x_2,x_3) \cdot \\ f_3(x_3,x_4)$, where $f_i(x_i, x_{i+1}) = xor(x_i,x_{i+1}) + 1$. Note that $\VIGnl$ of $f$ is a full graph. $\VIGnm$ and LyMPuS for $f$ are presented in Tables \ref{tab:Ex:fe1:VIGnm} and \ref{tab:Ex:fe1:LiPMoS}, respectively. For each variable $x_g$, LyMPuS stores $2^{|C_{\VIGnm}^g|}$ hyperplanes to determine the output of $b(x_g, C_{\VIG}^g(\vec{x}),\vec{x})$ for any $\vec{x}$ (Table \ref{tab:Ex:fe1:LiPMoS}, the last column). For instance, $b(x_1, C_{\VIG}^1(\vec{x}),\vec{x})$ depends only on $x_2$, i.e., $b(x_1, \text{*0**},\vec{x}) = \{1\}$ and $b(x_1, \text{*1**},\vec{x}) = \{0\}$. In contrast, $b(x_2, C_{\VIG}^2(\vec{x}),\vec{x})$ depends on $x_1$ and $x_3$ and some of its outputs contain more than value, e.g., $b(x_2,\text{0*1*},\vec{x}) = \{0,1\}$ (in the 0*1* hyperplane, both values of $x_2$ yield the same fitness).\par
\endexample

From this point onward, unless stated otherwise, \textbf{we assume $\VIGnm$ and $\eVIGnm$ whenever we refer to $\VIG$ and $\eVIG$, respectively}. In the black-box scenario, $\VIG$ is not available but we can use $\eVIG$ instead and construct the \textit{empirical} LyMPuS (eLyMPuS), i.e., LyMPuS constructed for $\eVIG$. To determine any $b(x_g, C_{\eVIG}^g(\vec{x}),\vec{x})$, we have to compute $b(x_g,\cdot,\vec{x})$ which equals $b(x_g, C_{\VIG}^g(\vec{x}),\vec{x})$. Thus, even if $C_{\eVIG}^g$ is incomplete in a given moment, the value of $b(x_g,\cdot,\vec{x})$ will remain valid after $C_{\eVIG}^g$ is supplemented, i.e., no fitness evaluations spared for constructing eLyMPuS will be wasted.\par

Walsh-based models enable \textit{partial evaluations}, i.e., re-evaluating only a part of the model when only a few variables were modified. Analogously, LyMPuS allows comparing solutions that differ by one gene concerning only a part of their genotypes, i.e., $C_{\VIG}^g$. Therefore, we call this functionality \textit{partial comparisons}.\par

\subsection{Partial comparisons, guaranteed linkage discovery, and eLyMPuS construction}
\label{sec:LiPMos:construction}

Many surrogate-assisted optimizers (including those using Perfect Linear Surrogates) are divided into two main steps, i.e., the surrogate-building step, and the surrogate-using (pure optimization) step \cite{ellGomea,verelSparseWalshModels}. However, in some cases, to find a solution of satisfactory quality, the surrogate does not have to model the entire solution space perfectly. 
%In contrast, in other cases, the model may have to be perfect. 
Therefore, the aforementioned separation of the surrogate-building and surrogate-using steps can be considered inflexible and less effective than updating the surrogate iteratively as a part of the optimization process. The same observation applies to modeling variable dependencies \cite{linkLearningDetermined}.\par

eLyMPuS is dedicated to black-box scenarios and is constructed on the fly. It will evaluate $f(\vec{x})$ only in three situations: to verify its correctness, to discover missing variable dependencies, or if $b(x_g, C_{\eVIG}^g(\vec{x}),\vec{x})$ was not computed for a given $C_{\eVIG}^g(\vec{x})$ yet.

\begin{algorithm}
    \small
	\caption{eLyMPuS Initialization}
	\begin{algorithmic}[1]
        \Procedure{Initialize}{$eLyMPuS$} \label{alg:LiPMos:init}
            \State  $eLyMPuS.\eVIG \gets $  empty; \label{alg:LiPMos:init:eVIG}
            \For{\textbf{each} $x_g$ \textbf{in} $\{x_1,...,x_n\}$}
                \State $eLyMPuS.x_gPairs(\vec{x},b^*_g(\vec{x})) \gets $ empty; \label{alg:LiPMos:init:xiPairs}
            \EndFor 
		\EndProcedure
    \end{algorithmic}
	\label{alg:LiPMosInit}
\end{algorithm}

Pseudocode \ref{alg:LiPMosInit} presents the eLyMPuS initialization procedure executed once at the start of the eLyMPuS-using optimizer. For each $x_g$, eLyMPuS maintains a set of initially empty pairs $\bigl(\vec{x}$,$b^*_g(\vec{x})\bigr)$, where $b^*_g(\vec{x}) = b(x_g, \cdot,\vec{x})$.% (we will use this shorter notation for the sake of readability).

\begin{algorithm}
    \small
	\caption{LyMPuS partial comparison procedure}
	\begin{algorithmic}[1]
        \Function{PartialComparison}{$x_g$, $\vec{x_{chk}}$, $eLyMPuS$, $Verify$}
            \State $Pair \gets $ FindPair($x_g$, $\vec{x_{chk}}$, $eLyMPuS$); \label{alg:LiPMos:getResult:pairSearch} \Comment{line \ref{alg:LiPMos:findPair}}
            %\State  $\eVIGnm \gets LiPMoS.\eVIGnm$;
            %\For{\textbf{each} $Pair(\vec{x},x^*_g)$ \textbf{in} $LiPMoS.x_gPairs(\vec{x},x^*_g)$}
                %\If{$C_{\eVIGnm}^g(Pair.\vec{x}) = C_{\eVIGnm}^g(\vec{x_{check}})$}
                \If{$Pair \neq $ empty}
                    \If{$Verify$ = true}
                        \State $b^*_{g,chk} \gets $  Compute\_$b^*_g$($x_g$, $x_{chk}$);  \label{alg:LiPMos:getResult:checkCorrectness} \Comment{line \ref{alg:LiPMos:compXg}}
                        \If{$Pair.b^*_g \neq b^*_{g,chk}$} \label{alg:LiPMos:getResult:missingLinkage}
                            \State RecursiveLL($Pair.\vec{x}$, $x_{chk}$, $x_g$, $eLyMPuS$); \label{alg:LiPMos:getResult:recursiveLL}
                            \State \parbox[t]{2em}{AddNewPair$\bigl(eLyMPuS.x_gPairs(\vec{x},b^*_g(\vec{x}))$, \\ \textcolor{white}{TAB}$Pair(\vec{x_{chk}},b^*_{g,chk})\bigr)$;\strut}
                            \State \Return $b^*_{g,chk}$;
                        \EndIf
                    \EndIf
                    \State \Return $Pair.b^*_g$;  \label{alg:LiPMos:getResult:resultSimple}
                \EndIf
            %\EndFor

            \State $b^*_{g,res} \gets $  Compute\_$b^*_g$($x_g$, $\vec{x_{chk}}$);    \label{alg:LiPMos:getResult:computeXgStar} \Comment{line \ref{alg:LiPMos:compXg}}
            \State AddNewPair$\bigl(eLyMPuS.x_gPairs(\vec{x},b^*_g(\vec{x}))$,$Pair(\vec{x_{chk}},b^*_{g,res})\bigr)$;\label{alg:LiPMos:getResult:supplementList}
            \State \Return $b^*_{g,res}$;
        \EndFunction
        \newline
        \Function{FindPair}{$x_g$, $\vec{x_{chk}}$, $eLyMPuS$} \label{alg:LiPMos:findPair}
            \State  $\eVIG \gets eLyMPuS.\eVIG$;
            \For{\textbf{each} $Pair\bigl(\vec{x},b^*_g(\vec{x})\bigr)$ \textbf{in} $eLyMPuS.x_gPairs\bigl(\vec{x},b^*_g(\vec{x})\bigr)$}
                \If{$C_{\eVIG}^g(Pair.\vec{x}) = C_{\eVIG}^g(\vec{x_{chk}})$}
                    \State \Return $Pair$;
                \EndIf
            \EndFor
            \State \Return empty;
        \EndFunction
        \newline
        \Function{Compute\_$b^*_g$}{$x_g$, $\vec{x_{chk}}$} \label{alg:LiPMos:compXg}
            \State $fit \gets $ fit($\vec{x_{chk}}$);
            \State $fit_g \gets $ fit($\vec{x^g_{chk}}$);

            \If{$fit < fit_g$}
                 $b^*_g \gets \{\vec{x^g_{chk}}[g]\}$
            \Else
                \If{$fit > fit_g$}
                     $b^*_g \gets \{\vec{x_{chk}}[g]\}$
                \Else
                    \text{ } $b^*_g \gets \{0, 1\}$
                \EndIf
            \EndIf

            \State \Return $b^*_g$;
        \EndFunction
    \end{algorithmic}
	\label{alg:LiPMos}
\end{algorithm}

In the black-box scenarios, we never know if $C_{\eVIG}^g = C_{\VIG}^g$. Therefore, for each $x_g$, we store a list of pairs $\bigl(\vec{x},b_g^*(\vec{x})\bigr)$. The list is supplemented in subsequent partial comparison executions for such $x_g$ and $\vec{x_{chk}}$ that $C_{\eVIG}^g(\vec{x_{chk}})$ does not exist in the list for a given $x_g$. If new dependencies are found during the run and $C_{\eVIG}^g$ is supplemented, all pairs in the list remain useful, i.e., the already paid cost of building the model is never wasted.\par

During partial comparison (Pseudocode \ref{alg:LiPMos}) for $\vec{x_{chk}}$ and $x_g$, eLyMPuS searches the list of pairs $(\vec{x},b^*_g(\vec{x}))$ (line \ref{alg:LiPMos:getResult:pairSearch}). If it finds a pair such that $C_{\eVIG}^g(Pair.\vec{x}) = C_{\eVIG}^g(\vec{x_{chk}})$, i.e., for all variables dependent on $x_g$ in $\VIG$, their values in $Pair.\vec{x}$ and $\vec{x_{chk}}$ are equal, then $(Pair.b^*_g)$ is a result (line \ref{alg:LiPMos:getResult:resultSimple}). If the appropriate pair does not exist, $b^*_g(\vec{x_{chk}})$ is computed and a new pair is added to the list (lines \ref{alg:LiPMos:getResult:computeXgStar}-\ref{alg:LiPMos:getResult:supplementList}).\par

eLyMPuS uses $\eVIG$ that may be incomplete. Therefore, sometimes it should verify if linkage is not missing. In such a case, $b^*_g(\vec{x_{chk}})$ is computed (line \ref{alg:LiPMos:getResult:checkCorrectness}). If $\eVIG$ contains all dependencies for $x_g$, i.e., $C^g_{\eVIG} = C^g_{\VIG}$, then it is guaranteed that $b^*_g(\vec{x_{chk}}) = b(x_g, C_{\eVIG}^g(\vec{x_{chk}}),Pair.\vec{x})$. In the opposite situation, the results may differ (line \ref{alg:LiPMos:getResult:missingLinkage}). Then, we \textit{\textbf{know}} that differing results are caused by the missing dependency between $x_g$ and at least one variable that differs in $\vec{x_{chk}}$ and $Pair.\vec{x}$.\par

\begin{algorithm}
    \small
	\caption{Guaranteed Linkage Discovery with Recursive LL}
	\begin{algorithmic}[1]
        \Procedure{RecursiveLL}{$\vec{x_{s1}}$, $\vec{x_{s2}}$, $x_g$, $eLyMPuS$}
            \State $\eVIG \gets eLyMPuS.\eVIG$;
            \State $GenesToConsider \gets $ Diff($\vec{x_{s1}}$, $\vec{x_{s2}}$) $\text{ }\setminus \text{ } x_g$; \label{alg:RecursiveLL:CInit}

            \State RecForGroup($\vec{x_{s1}}$, $\vec{x_{s2}}$, $x_g$, $GenesToConsider$, $eLyMPuS$);
		\EndProcedure
        \newline
        \Procedure{RecForGroup}{$\vec{x_{s1}}$, $\vec{x_{s2}}$, $x_g$, $GroupToDiv$, $eLyMPuS$}
            \If{$|GroupToDiv| = 1$} \label{alg:RecursiveLL:vigUpStart}
                \State $eLyMPuS.\eVIG[x_g][GroupToDiv[1]] \gets 1$;
                \State $eLyMPuS.\eVIG[GroupToDiv[1]][x_g] \gets 1$; \label{alg:RecursiveLL:vigUpEnd}
                \State \Return;
            \EndIf
            \State $b^*_{g,s2} \gets $  Compute\_$b^*_g$($x_g$, $\vec{x_{s2}}$); \Comment{Pseudocode \ref{alg:LiPMos}, line \ref{alg:LiPMos:compXg}}
            \State $gr_A, gr_B  \gets $ RandDivideInTwoEqualGroups ($GroupToDiv$); \label{alg:RecursiveLL:twoGroups}
            \State $\vec{x'_{s1}} \gets \vec{x_{s1}}$;
            $\vec{x'_{s1}} \gets_{gr_A} \vec{x_{s2}}$; \label{alg:RecursiveLL:xs1Prim}
            %\For{\textbf{each} $x_i$ \textbf{in} $gr_A$}
             %   \State $\vec{x'_{s1}}[i] \gets \vec{x_{s2}}[i]$;
            %\EndFor
            \State $x^*_{g,s'1} \gets $  Compute\_$b^*_g$($x_g$, $\vec{x'_{s1}}$); \label{alg:RecursiveLL:xgStar}

            \If{$b^*_{g,s'1} = b^*_{g,s2}$}
                \State RecForGroup($\vec{x_{s1}}$, $\vec{x'_{s1}}$, $x_g$, $gr_A$, $eLyMPuS$);  \label{alg:RecursiveLL:divide}
            \Else
                \State RecForGroup($\vec{x'_{s1}}$, $\vec{x_{s2}}$, $x_g$, $gr_B$, $eLyMPuS$); \label{alg:RecursiveLL:supplement}
            \EndIf
        \EndProcedure
    \end{algorithmic}
	\label{alg:RecursiveLL}
\end{algorithm}

The above procedure finds two solutions ($Pair.\vec{x}$ and $\vec{x_{chk}}$) for which $b^*_g(\vec{x_{chk}}) \neq b^*_g(Pair.\vec{\vec{x}})$, although $C_{\eVIG}^g(\vec{x_{chk}}) = C_{\eVIG}^g(Pair.\vec{x})$. Thus, $C_{\eVIG}^g \subset C_{\VIG}^g$. For the sake of clarity, in Pseudocode \ref{alg:RecursiveLL}, we denote these two solutions as $\vec{x_{s1}}$, $\vec{x_{s2}}$. Let $\mathcal{S}=\{\vec{x_{0}},...,\vec{x_{|C|}}\}$ be a sequence of solutions such that:
\begin{enumerate}
    \item $C = \text{Diff}(\vec{x_{s1}}, \vec{x_{s2}}) \setminus x_g$, where $\text{Diff}(\vec{x_{s1}}, \vec{x_{s2}})$ is the set of all variables that differ in $\vec{x_{s1}}$, $\vec{x_{s2}}$ (note that $C_{\eVIG}^g(\vec{x_{s1}}) = C_{\eVIG}^g(\vec{x_{s2}})$.
    \item $\vec{x_0} = \vec{x_{s1}}$
    \item For each $x_{c_i} \in C$, considered in any order, $\vec{x_{i}} = \vec{x_{i-1}} \xleftarrow[x_{c_i}]{}~\vec{x_{s2}}$, where $\vec{x_a} \xleftarrow[m]{} \vec{x_b}$ denotes copying variable values marked by a mask $m$ from $\vec{x_b}$ to $\vec{x_a}$. Note that $\vec{x_{i}} = \vec{x_{i-1}}^{x_{c_i}}$, i.e., $\vec{x_{i-1}}$ with $x_{c_i}$ flipped.
\end{enumerate}

\begin{thm}
    \label{thm:certainDiscovery}
        Sequence $\mathcal{S}$ contains two consecutive solutions $\vec{x_{i-1}}$, $\vec{x_{i}}$ such that $\vec{x_{i-1}}$, $\vec{x_{i-1}}^g$, $\vec{x_{i-1}}^{c_i}$, $\vec{x_{i-1}}^{g,c_i}$ fulfill C1, C2 or C3 clause.
    \begin{proof}
        The definition of $\mathcal{S}$, states that $b^*_g(\vec{x_0}) = b^*_g(\vec{x_{s1}}) \neq b^*_g(\vec{x_{s2}}) = b^*_g(\vec{x_{|C|}})$. All neighbouring solutions in $\mathcal{S}$ differ by one gene. Thus, $\mathcal{S}$ contains consecutive solutions, $\vec{x_{i-1}}$, $\vec{x_{i}}$ such that $b^*_g(\vec{x_{i-1}}) \neq b^*_g(\vec{x_i})$, therefore, $f(\vec{x_{i-1}}) R_{i-1} f(\vec{x_{i-1}}^g)$, $f(\vec{x_{i}}) R_{i} f(\vec{x_{i}}^g)$, and $R_{i-1} \neq R_i$. Since $\vec{x_i} = \vec{x_{i-1}}^{c_i}$, then $f(\vec{x_{i-1}}^{c_i}) R_{i} f(\vec{x_{i}}^{g,c_i})$, which proves that using $\vec{x_{i-1}}$, $x_g$, and $x_{c_i}$ must fulfil one of C1-C3 clauses (for each clause, $h=c_i$).
    \end{proof}
\end{thm}
The above theorem proves that when \textproc{PartialComparison} finds missing linkage for solutions $\vec{x_{chk}}$ and $Pair.\vec{x}$, then we are \textit{guaranteed} to find one dependency that is missing in $eLyMPuS.\eVIG$. Moreover, we are only guaranteed to fill one of the C1-C3 clauses (not the C4-C6). Thus, we can distinguish directional and symmetrical dependencies. The dependency will be discovered in no more than $|C|$ steps. If we perform this operation recursively and split $C$ into two equal (or almost equal) groups in each step, we are guaranteed to find one missing dependency in $\lceil \log_2(|C|) \rceil$ steps.\par

Pseudocode \ref{alg:RecursiveLL} presents the recursive linkage discovery version. Initially, we consider all genes in $C$ (line \ref{alg:RecursiveLL:CInit}). In each step, we split $GenesToConsider$ into two groups ($gr_A, gr_B$) of equal size (or sizes differing by one if $|GenesToConsider|$ is an odd number). We create $\vec{x'_{s1}}$, a copy of $\vec{x_{s1}}$, and the genes from $\vec{x_{s2}}$ marked by $gr_A$. If $b^*_g(\vec{x'_{s1}}) = b^*_g(\vec{x'_{s2}})$, we split $gr_A$ (line \ref{alg:RecursiveLL:divide}). Otherwise, we split $gr_B$ for solutions $\vec{x'_{s1}}$ and $\vec{x_{s2}}$ (line \ref{alg:RecursiveLL:supplement}). When $|GenesToConsider| = 1$, we are guaranteed that we have found two solutions that differ by one gene for which flipping $x_g$ yields different $b^*_g(\vec{x})$.\par

 \begin{table}  
    \caption{$\eVIG$ discovered by eLyMPuS in Example \ref{ex:recursiveLL} (missing dependencies are marked in \textcolor{red}{\textbf{red}})}
    \label{tab:Ex:fe2:eVIGnm}
    \scriptsize
    \centering 
       \begin{tabular}{l|cccccccc}
              & $x_1$  & $x_2$  & $x_3$ & $x_4$ & $x_5$ & $x_6$ & $x_7$ & $x_8$ \\
              \hline
            $x_1$ & X & 1 & \textcolor{red}{\textbf{0}} & \textcolor{red}{\textbf{0}} & 0 & 0 & 0 & 0 \\
            $x_2$ & 1 & X & \textcolor{red}{\textbf{0}} & \textcolor{red}{\textbf{0}} & 0 & 0 & 0 & 0 \\
            $x_3$ & \textcolor{red}{\textbf{0}} & \textcolor{red}{\textbf{0}} & X & 1 & \textcolor{red}{\textbf{0}} & \textcolor{red}{\textbf{0}} & 0 & 0 \\
            $x_4$ & \textcolor{red}{\textbf{0}} & \textcolor{red}{\textbf{0}} & 1 & X & \textcolor{red}{\textbf{0}} & \textcolor{red}{\textbf{0}} & 0 & 0 \\
            $x_5$ & 0 & 0 & \textcolor{red}{\textbf{0}} & \textcolor{red}{\textbf{0}} & X & 1 & \textcolor{red}{\textbf{0}} & \textcolor{red}{\textbf{0}} \\
            $x_6$ & 0 & 0 & \textcolor{red}{\textbf{0}} & \textcolor{red}{\textbf{0}} & 1 & X & \textcolor{red}{\textbf{0}} & \textcolor{red}{\textbf{0}} \\
            $x_7$ & 0 & 0 & 0 & 0 & \textcolor{red}{\textbf{0}} & \textcolor{red}{\textbf{0}} & X & 1 \\
            $x_8$ & 0 & 0 & 0 & 0 & \textcolor{red}{\textbf{0}} & \textcolor{red}{\textbf{0}} & 1 & X 
        \end{tabular}
\end{table}
 \begin{table}  
 \caption{eLyMPuS for Example \ref{ex:recursiveLL}}
    \label{tab:Ex:fe2:LiPMoS}
    \scriptsize
    \centering 
       \begin{tabular}{l|cccc}
            $x_g$ & $C_{\eVIG}^g$ & $Pair.\vec{x}$ & $C_{\eVIG}^g(Pair.\vec{x})$  & $b^*_g(Pair.\vec{x})$ \\
              \hline
            $x_1$ & $\{x_2\}$ & 01 10 10 01 & *1 ** ** ** & $\{0\}$ \\
                  &           & 10 10 10 01 & *0 ** ** ** & $\{1\}$ \\
                  \hline
            $x_2$ & $\{x_1\}$ & 10 10 10 01 & 1* ** ** ** & $\{0\}$ \\
                  \hline
            $x_3$ & $\{x_4\}$ & 11 11 00 11 & ** *1 ** ** & $\{1\}$ \\
                  &           & 00 00 11 00 & ** *0 ** ** & $\{0\}$ \\
                  \hline
            $x_4$ & $\{x_3\}$ & 01 10 01 10 & ** 1* ** ** & $\{0\}$ \\
                  &           & 10 01 01 10 & ** 0* ** ** & $\{1\}$ \\
                  \hline
            $x_5$ & $\{x_6\}$ & 01 10 10 01 & ** ** *0 ** & $\{1\}$ \\
                  \hline
            $x_6$ & $\{x_5\}$ & 01 10 10 01 & ** ** 1* ** & $\{0\}$ \\
                  \hline
            $x_7$ & $\{x_8\}$ & 01 10 10 01 & ** ** ** *1 & $\{0\}$ \\
                  \hline
            $x_8$ & $\{x_7\}$ & 01 10 10 01 & ** ** ** 0* & $\{1\}$ \\
                  \hline
        \end{tabular}
\end{table}

\example \label{ex:recursiveLL}
Consider $f(x_1,...,x_8) = f_1(x_1,...,x_4) + f_2(x_3,...,x_6) + f_3(x_5,...,x_8)$, where $f_i = bim_4(x_{2i-1},...,x_{2i+2})$. We use eLyMPuS from Table \ref{tab:Ex:fe2:LiPMoS} to locally optimize $\vec{x} = [11\text{ }10\text{ }01\text{ }01]$; $f(\vec{x}) = f_1(1110) + f_2(1001) + f_3(0101) = 0 + 1 + 1 = 2$. We try to perform bitflips in the gene order. We use partial comparisons and for each such operation, we verify if the eLyMPuS result is correct. For $x_1$, $C_{\eVIG}^1(Pair.\vec{01 10 10 01}) = C_{\eVIG}^1(\vec{x}) = $ [*1 ** ** **], and eLyMPuS returns $\{0\}$. We compute $b^*_1(\vec{x}) = \{0\}$ that confirms eLyMPuS answer (Pseudocode \ref{alg:LiPMos}, line \ref{alg:LiPMos:getResult:checkCorrectness}). Solution $\vec{x}$ is modified to $\vec{x}^1 = [\textbf{\textit{0}}1\text{ }10\text{ }01\text{ }01]$.\par

Next, we try to flip $x_2$ in $\vec{x}^1$. However, $C_{\eVIG}^2(\vec{x}^1) = $ 0* ** ** **, and there is no solution in the list of pairs $\bigl(\vec{x},b_2^*(\vec{x})\bigr)$ that would fit to it. Therefore, we compute $b^*_2(\vec{x}^1) = \{1\}$, and add the new pair $\bigl(\vec{x}^1, b^*_2(\vec{x}^1)\bigr)$ to the list of pairs for $x_2$ (Pseudocode \ref{alg:LiPMos}, line \ref{alg:LiPMos:getResult:supplementList}).\par

In the third step, we try to modify $x_3$. eLyMPuS answer, $\{0\}$, is incorrect because $b^*_3(\vec{x}^1) = \{1\}$. Thus, we execute \textproc{RecursiveLL} (Pseudocode \ref{alg:LiPMos}, line \ref{alg:LiPMos:getResult:recursiveLL}), and $C = \text{Diff}(\vec{00 00 11 00}, \vec{x}^1) \setminus x_g = \{x_2, x_5, x_8\}$ (Pseudocode \ref{alg:RecursiveLL}, line \ref{alg:RecursiveLL:CInit}). At least one variable in $C$ depends on $x_3$. We randomly split $C$ into two subsets (Pseudocode \ref{alg:RecursiveLL}, line \ref{alg:RecursiveLL:twoGroups}). Assume $gr_A = \{x_5\}$ and $gr_B = \{x_2, x_8\}$, then $x'_{s1} = 00\text{ }00\text{ }\textbf{\textit{0}}1\text{ }00$ and $b^*_3(\vec{x'_{s1}}) = \{0\} \neq b^*_3(\vec{x_{s2}}) = b^*_3(\vec{x}^1) = \{1\}$ (lines \ref{alg:RecursiveLL:xs1Prim} and \ref{alg:RecursiveLL:xgStar}). Thus, we will split $gr_B = \{x_2, x_8\}$ for solutions $\vec{x'_{s1}}$ and $\vec{x_{s2}} = \vec{x}^1$ (line \ref{alg:RecursiveLL:supplement}). Assume that after split we receive $gr_A = \{x_2\}$ and $gr_B = \{x_8\}$, then $\vec{x''_{s1}} = 0\textbf{\textit{1}}\text{ }00\text{ }01\text{ }00$ and $b^*_3(\vec{x''_{s1}}) = \{1\} = b^*_3(\vec{x_{s2}}) = b^*_3(\vec{x}^1)$. Thus, we try to split $gr_A = \{x_2\}$ for solutions $\vec{x'_{s1}}$ and $\vec{x''_{s1}}$ (line \ref{alg:RecursiveLL:divide}). Since $|gr_A| = 1$, $x_3$ depends on variable in $gr_A$, i.e., $x_2$ (Pseudocode \ref{alg:LiPMos}, lines \ref{alg:RecursiveLL:vigUpStart}-\ref{alg:RecursiveLL:vigUpEnd}).\par
\endexample

In Example \ref{ex:recursiveLL}, the last execution of \textproc{RecForGroup} took place for $\vec{x'_{s1}}$ and $\vec{x''_{s1}}$ that differ only by $x_2$ and $b^*_3(\vec{x'_{s1}}) \neq b^*_3(\vec{x''_{s1}})$. Thus, these can be two consecutive solutions $\vec{x_{i-1}}$, $\vec{x_{i}}$ concerned in Theorem \ref{thm:certainDiscovery}.\par

Here, we focus on symmetrical dependencies. However, the above procedure allows for directional dependency discovery. Such a feature is beneficial because every symmetrical dependency is built from two directional ones. For problems with directional dependencies, discovering them enables constructing more precise $\eVIG$, which may lead to a significant increase in effectiveness \cite{2dled}.\par

Pseudocodes \ref{alg:LiPMos} and \ref{alg:RecursiveLL} aim to present the algorithm idea rather than reflect the actual source code. The fitness-based cost of \textproc{PartialComparison} is $0$ if $Verify$ = false or $1$ otherwise. In \textproc{RecursiveLL}, the cost of a single \textproc{RecForGroup} execution is 2 because we only have to compute $f(\vec{x'_{s1}})$ and $f(\vec{x'_{s1}}^g)$. Thus, the FFE cost of \textproc{RecursiveLL} equals $2 \lceil\log_2(|C|)\rceil$.

%\textcolor{red}{\textbf{Marcin: Perhaps it is good to underline here (in 2-3 sentences) that we are significantly different than any other recursive LL we know?}}

\subsection{First Improvement Hill Climber with eLyMPuS and the Circuit Check}
\label{sec:LNMSurr:surrLL}

eLyMPuS is suitable for any local search procedure in which, in one step, only one variable is modified. If result correctness is verified at every partial comparison execution, the FFE cost of the local search will remain unchanged. However, eLyMPuS may discover missing dependencies (and subsequently, dependencies in the low-cost procedure presented in the prior subsection). Thus, in such a setting, eLyMPuS will support a zero-cost missing linkage detection and a low-cost linkage discovery with guaranteed output.\par

\begin{algorithm}
    \small
	\caption{First Improvement Hill Climber using eLyMPuS}
	\begin{algorithmic}[1]
        \Function{FIHC-eLyMPuS}{$\vec{x_{chk}}$, $eLyMPuS$}
            \State $Verify \gets $ eLyMPuSVerifyStrategy(); \label{alg:FIHCLiPMos:verify}

            \State $Mods(\vec{x}, i, b^*_i) \gets $ empty;  \label{alg:FIHCLiPMos:mods:empty}
            \State $\pi_n \gets $ GenerateRandomPermutation(1,$n$);

            \Repeat
                \State  $modified \gets $ false;
                \For{\textbf{each} $i$ \textbf{in} $\pi_n$}
                    \State $b^*_i \gets $ PartialComparison($x_i$,$\vec{x_{chk}}$, $eLyMPuS$, $Verify$);
                    \If {$|b^*_i|$ = 1 \textbf{and} $b^*_i \neq \vec{x_{chk}}[i]$}
                        \State $Mods \gets Mods + (\vec{x_{chk}}, i, b^*_i)$; \label{alg:FIHCLiPMos:mods:add}
                        \State $\vec{x_{chk}}[i] \gets b^*_i$;
                        \State  $modified \gets $ true;
                        \If {CheckCircuits($Mods$, $eLyMPuS$) = true} \label{alg:FIHCLiPMos:CheckCircuits:exec}%\Comment{line \ref{alg:POLiPMos:CheckCircuits}}
                            \State $Mods(\vec{x}, i, b^*_i) \gets $ empty;
                        \EndIf
                    \EndIf                
                \EndFor
            \Until{$modified = $ true}

            \State \Return $\vec{x_{chk}}$;
		\EndFunction
    \end{algorithmic}
	\label{alg:FIHCLiPMos}
\end{algorithm}

The above eLyMPuS setting is attractive but does not enable the FFE cost decrease, which is a typical surrogate goal. Therefore, we propose the First Improvement Hill Climber using eLyMPuS (FIHC-eLyMPuS). At the beginning of each FIHC-eLyMPuS execution, we decide if eLyMPuS will verify missing linkage during this execution or not (Pseudocode \ref{alg:FIHCLiPMos}, line \ref{alg:FIHCLiPMos:verify}). We use a simple strategy denoted as $1/v^{th}$. After each FIHC-eLyMPuS iteration with verification, we count the number of $v$ subsequent FIHC-eLyMPuS iterations with verification (which may be interrupted by iterations without verification) in which no missing linkage was found. Then, the next FIHC-eLyMPuS iteration with verification will be held after the next $v+1$ iterations without verification. Thus, if $\eVIG$ is complete, then the verification will be held more and more rarely, which decreases the overall optimization cost because eLyMPuS without verification is free in terms of FFE. The $1/v^{th}$ strategy ensures that verification is held sometimes, which is important in black-box scenarios.%, we do not know if $\eVIG = \VIG$.\par

\begin{algorithm}
    \small
	\caption{Circuit-based guaranteed to detect missing linkage}
	\begin{algorithmic}[1]
    \Function{CheckCircuits}{$Mods(\vec{x}, i, b^*_i)$,$eLyMPuS$} \label{alg:FIHCLiPMos:CheckCircuits}
            \State $LastSolModTripple \gets Mods[|Mods|]$;
            \For{$mod = 1$ \textbf{to} $|Mods|-1$}
                \If{$LastSolModTripple.\vec{x} = Mods[mod].\vec{x}$} \label{alg:FIHCLiPMos:CheckCircuits:Circuit}
                    \newline
                    \Comment{\textbf{\textit{Entering this block we are} guaranteed \textit{to find missing linkage and one missing dependency}}}
                    \For{$j = mod$ \textbf{to} $|Mods|-1$} \label{alg:FIHCLiPMos:CheckCircuits:searchStart}
                        \State $g \gets Mods[j].i$;
                        \State $b^*_g \gets $  Compute\_$b^*_g$($x_g$, $Mods[j].\vec{x}$); \Comment{Pseudocode \ref{alg:LiPMos}}
                        \If{$b^*_g \neq Mods[j].b^*_i$} \label{alg:FIHCLiPMos:CheckCircuits:searchFound}
                            \State $CircMod \gets $ FindPair($x_g$, $Mods[j].\vec{x}$, $eLyMPuS$);
                            %\State RecursiveLL($Mods[j].\vec{x}$, $Pair.\vec{x}$, $x_g$, $LiPMoS$);
                            \State \parbox[t]{2em}{RecursiveLL($Mods[j].\vec{x}$,$CircMod.\vec{x}$, \\ \textcolor{white}{TAB}$x_g$,$eLyMPuS$);\strut} \label{alg:FIHCLiPMos:CheckCircuits:recursiveLL}
                            \State \parbox[t]{2em}{AddNewPair$\bigl(eLyMPuS.x_gPairs(\vec{x},b^*_g)$, \\ \textcolor{white}{TAB}$Pair(Mods[j].\vec{x}, b^*_g)\bigr)$;\strut}
                            \State \Return true;
                        \EndIf
                    \EndFor
                \EndIf
            \EndFor

            \State \Return false;
        \EndFunction
    \end{algorithmic}
	\label{alg:FIHCLiPMos:CheckCircuits}
\end{algorithm}

In some cases, FIHC-eLyMPuS can detect missing linkage as a part of its procedure without verification. Let $\mathcal{T}=\{\vec{x_{0}},...,\vec{x_{|T|}}\}$ be a sequence of solutions yielded during FIHC-eLyMPuS execution in subsequent steps (Pseudocode \ref{alg:FIHCLiPMos}, lines \ref{alg:FIHCLiPMos:mods:empty} and \ref{alg:FIHCLiPMos:mods:add}). If $\vec{x_{|T|}} = \vec{x_{t}}$ where $t \in \{0,...,|T|-1\}$, then at least one move between $\vec{x_{t}}$ and $\vec{x_{|T|}}$ was not an improving one, i.e., $f(\vec{x_w}) \geq f(\vec{x_{w+1}})$, where $w \in \{t,...,|T|-1\}$. Thus, at least one $b^*_{g(w)}(\vec{x_{w}}) \neq eLyMPuS(x_{g(w)}, \eVIG, \vec{x_w})$, where $x_g(w)$ is the variable flipped in $\vec{x_w}$ to yield $\vec{x_{w+1}}$. Therefore, after each step found improving, \textproc{CheckCircuits} is executed (line \ref{alg:FIHCLiPMos:CheckCircuits:exec}).\par

\textproc{CheckCircuits} (Pseudocode \ref{alg:FIHCLiPMos:CheckCircuits}) checks if $\vec{x_{|T|}} = \vec{x_{t}}$. If so, then checks subsequent $b^*_{g(w)}(\vec{x_{w}})$ until finding the incorrect move (lines \ref{alg:FIHCLiPMos:CheckCircuits:searchStart}-\ref{alg:FIHCLiPMos:CheckCircuits:searchFound}) and executes \textproc{RecursiveLL} for it (line \ref{alg:FIHCLiPMos:CheckCircuits:recursiveLL}).\par

\example \label{ex:CheckCircuits}

Consider $f(x_1,...,x_8)$, $\vec{x}$ and eLyMPuS (Table \ref{tab:Ex:fe2:LiPMoS}) from Example \ref{ex:recursiveLL}. We use partial comparisons to optimize $\vec{x} = [11\text{ }10\text{ }01\text{ }01]$, but we do not verify the correctness of the partial comparison, i.e., $Verify$ = false. Genes are considered in the gene order. 
%The eLyMPuS-based decisions will be as follows. 
In the first iteration, FIHC-eLyMPuS will flip $x_1$, $x_3$, and $x_4$ (for $x_2$, $x_5$, and $x_6$ the lists of $\bigl(\vec{x},b_g^*(\vec{x})\bigr)$ pairs will be supplemented). In the second iteration, FIHC-eLyMPuS will flip $x_3$ and $x_4$. Thus, $\mathcal{T}=\{\vec{x_{0}} = 11\text{ }10\text{ }01\text{ }01, \textcolor{red}{\vec{x_{1}} = \textbf{\textit{0}}1\text{ }10\text{ }01\text{ }01}, \vec{x_{2}} = 01\text{ }\textbf{\textit{0}}0\text{ }01\text{ }01, \vec{x_{3}} = 01\text{ }0\textbf{\textit{1}}\text{ }01\text{ }01, \vec{x_{4}} = 01\text{ }\textit{\textbf{1}}1\text{ }01\text{ }01, \textcolor{red}{\vec{x_{5}} = 01\text{ }1\textbf{\textit{0}}\text{ }01\text{ }01}\}$, $\vec{x_1} = \vec{x_5}$ and one of the modifications for $\vec{x_1}$, $\vec{x_2}$, $\vec{x_3}$, or $\vec{x_4}$ was not an improving move. Indeed, $f(\vec{x_1}) > f(\vec{x_2} = \vec{x_1^3})$ and \textproc{RecursiveLL} will discover a dependency for $x_3$ and supplement $eLyMPuS.\eVIG$.\par
\endexample

%Note that searching for a circuit in $\mathcal{T}$ does not yield any fitness evaluations. If a circuit is found, then eLyMPuS is guaranteed to detect missing linkage in no more than $|\mathcal{T}|$ steps (and FFE) and discover one missing dependency using \textproc{RecurisveLL}.\par

%Searching for a circuit in $\mathcal{T}$ does not yield FFE. If a circuit is found, then eLyMPuS is guaranteed to detect missing linkage in no more than $|\mathcal{T}|$ steps (and FFE) and discover one missing dependency using \textproc{RecurisveLL}.\par

Detecting circuit in $\mathcal{T}$ does not yield FFE. If it is found, eLyMPuS is guaranteed to detect missing linkage in no more than $|\mathcal{T}|$ steps (and FFE) and discover one missing dependency using \textproc{RecurisveLL}.\par

\section{Recursive linkage learning via Partition Crossover in black-box scenarios}
\label{sec:PXrecursiveLL}

Originally, PX was proposed to use $\VIGnl$ in gray-box scenarios \cite{pxForBinary}. Recently, it was shown that PX applies to $\eVIGnm$ and black-box optimization as well \cite{dgga}. In such a setting, if a given PX mask is complete, i.e., it equals one of the masks that would be yielded for $\VIGnm$, then $\bigl(f(\vec{x_{s1}}) R_{ine} f(\vec{x_{o1}})$ \textbf{and} $f(\vec{x_{s2}}) \neg R_{ine} f(\vec{x_{o2}})\bigr)$ \textbf{or} $\bigl(f(\vec{x_{s1}}) = f(\vec{x_{o1}})$ \textbf{and} $f(\vec{x_{s2}}) = f(\vec{x_{o2}})\bigr)$, where $R_{ine} = \{>,<\}$, $\vec{x_{s1}}$ and $\vec{x_{s2}}$ are the mixed solutions, $\vec{x_{o1}} = \vec{x_{s1}} \xleftarrow[mPX]{} \vec{x_{s2}}$, $\vec{x_{o2}} = \vec{x_{s2}} \xleftarrow[mPX]{} \vec{x_{s1}}$, and $mPX$ is a PX mask. Otherwise, the mask is incomplete, and we are \textit{guaranteed} that at least one missing dependency exists between a gene from $mPX$ and the rest of the genotype \cite{dgga}. In the original proposition, after such missing linkage detection, all pairs joining each gene from $mPX$ and outside of it are the subject to the non-monotonicity check \cite{dgga}. Such a procedure is expensive, and it does not guarantee any output. We propose \textit{PX with Recursive Linkage Learning} (PXrLL) that guarantees finding $x_g$ on which (the incomplete) $mPX$ depends in $log_2(|\text{Diff}(\vec{x_{s1}}, \vec{x_{s2}}) \setminus mPX|)$ steps. We also propose a recursive procedure to find which gene from $mPX$ depends on $x_g$ but it does not guarantee to find the dependency. Since the procedure of PXrLL is based on the same ideas as presented in Section \ref{sec:LNMSurr}, its detailed presentation, pseudocodes and examples can be found in Section S-III (supplementary material).\par

\section{Pyramid-based Optimizer using eLyMPuS}
\label{sec:POLiPMoS}

FIHC-eLyMPuS can be utilized in any optimizer as a local search algorithm. We introduce it to P3-FIHCwLL and LT-GOMEA-FIHCwLL \cite{FIHCwLL} where it replaces FIHCwLL. However, to fully utilize the potential of eLyMPuS and other proposed mechanisms, we propose the Optimizer using eLyMPuS (OLyMPuS). OLyMPuS is a parameter-less optimizer that uses a pyramid-like population management, PXrLL and the ILS-like solution improvement step supported by eLyMPuS. Below, we refer to Pseudocode S-3 (supplementary material) that presents the details of OLyMPuS.  \par

In each iteration, \textproc{PXlinkDiscovery} is executed in which two individuals are created randomly and mixed using PX and $\eVIG$. The only purpose of this operation is linkage discovery. Note that the cost of this operation (4 FFE) is negligible but sometimes it may discover dependencies.\par

In the main part of each iteration, OLyMPuS randomly creates a new individual $\vec{x_{climb}}$, optimizes it with FIHC-eLyMPuS, and inserts it into the first level of the pyramid (Pseudocode S-3, lines 6-8). 
%(lines \ref{alg:POLiPMoS:newInd}-\ref{alg:POLiPMoS:newIndinsert}). 
Then, $\vec{x_{climb}}$ is the subject of the ILS-like optimization step, i.e., for each $x_g$ (considered in a random order), genes marked by $C_{\eVIG}^g \cup x_g$ are randomly reinitialized, FIHC-eLyMPuS optimizes the perturbed $\vec{x'_{climb}}$ 
%(lines \ref{alg:POLiPMoS:ils:perturb}-\ref{alg:POLiPMoS:ils:fihc}). 
(lines 15-16). If the fitness of the resulting $\vec{x'_{climb}}$ is not worse than for $\vec{x_{climb}}$, it replaces $\vec{x_{climb}}$. If \textproc{ILSlikeOpt} improves $\vec{x_{climb}}$, its improved copy is added to the next level of the pyramid 
%(line \ref{alg:POLiPMoS:newILSinsert}). 
(line 19). 
Finally, $\vec{x_{climb}}$ is mixed with subsequent levels of the pyramid using PXrLL 
%(line \ref{alg:POLiPMoS:CLimbUp}).\par
(line 22).\par

\section{Experiments}
\label{sec:expMain}
%\textcolor{red}{\textbf{Research Questions:}}\\
%RQ1. Can the limited surrogate speedup solution comparison for %problems that are not considered as expensive-in-evaluation?

\subsection{Experiment setup and considered optimizers}
\label{sec:expMain:setup}

OLyMPuS is an optimizer designed to leverage eLyMPuS's advantages. Additionally, to verify the effectiveness of eLyMPuS, we introduce it into P3-FIHCwLL and LT-GOMEA-FIHCwLL \cite{FIHCwLL} by replacing the FIHCwLL mechanism with eLyMPuS. Therefore, we consider P3-FIHCwLL, LT-GOMEA-FIHCwLL, and their original versions, i.e., P3 and LT-GOMEA, as baseline optimizers. Due to space limitations, LT-GOMEA results are reported only in the supplementary material. Each experiment is repeated 30 times.\par

The computation budget is set on $2\cdot 10^7$ FFE and $12$ hours. If any experiment exceeds at least one of these constraints, the experiment is stopped. Such a stop condition is designed to limit the advantage of OLyMPuS, which consumes the FFE-based budget very slowly due to leveraging the advantages of eLyMPuS.\par

We consider a set of well-known benchmark problems that includes: bimodal-10 (Bim10), noised bimodal-10 (nBim10), and standard deceptive functions ($k=5$, Dec5). We consider concatenations of these functions and cyclic traps where functions share variables (e.g., in Bim10o2, Bim10 functions share 2 variables with each of their neighbours). We also consider NK-fitness landscapes (nkLand) \cite{P3Original,nkLandscapes}, Ising Spin Glasses (ISG) \cite{isg}, max3sat (m3s) \cite{transTokBounded,P3Original}, mk fitness landscapes using Bim10 and Dec3  \cite{mkLandDirk,grayWhitley}. The detailed definitions can be found in Section S-V (supplementary material).\par

The implementation of all optimizers and problems was joined in one C++ project and is available at GitHub\footnote{\url{https://github.com/przewooz/OLyMPuS}}. To assure that the time-based stop condition is reliable, all experiments were single-threaded and executed on the PowerEdge R6525 Dell Server AMD 2 Epyc 7H12 2.6 GHz 1TB RAM. Each experiment was repeated 30 times, and the experiment guiding software kept the number of computing processes constant and lower than the number of physical processor cores.\par

%To show the meaning of POLiPMoS mechanisms we consider it in several versions, i.e., standard, without linkage learning in PXrLL, with brutal linkage learning after finding missing linkage in PXrLL (after detecting the missing linkage it performs the non-monotonicity check for each pair created by a variable from the incomplete mask and the rest of the genotype), without the ILS-like step and without the PX-based climbing. Finally, we consider P3 and LT-GOMEA in their original versions, FIHCwLL versions and using eLiPMoS.\par

Optimizers capable of finding the optimal solution for larger instances were found more effective. If more optimizers scaled similarly, then the statistical significance of consumed FFE-budget was considered (depending on the instance group, we have employed the Sign test or the unpaired Wilcoxon test, both with the 5\% significance level).

\subsection{Main results}
\label{sec:expMain:results}

\begin{table*}[]
\caption{OLyMPuS -- detailed results for chosen instances (median values reported). Full results: Tables S-II and S-II (supplementary material).}
    \label{tab:polipmos:detailed}
    \scriptsize
    \centering 
\begin{tabular}{l|rlc|llll|llll|l|rrrr}

        &      &  & \textbf{Dep.} &  & & & & & & & &  & \multicolumn{4}{c}{\textbf{Linkage discovery (ratio [\%])}}     \\
        
        &     \textbf{Succ} & \textbf{med. FFE until best} & \textbf{disc.} & \multicolumn{4}{c|}{\textbf{FIHC-eLyMPuS FFE cost}} & \multicolumn{4}{c|}{\textbf{PXrLL FFE cost}} & \textbf{InitPX} & \multicolumn{2}{c}{\textbf{eLyMPuS}} &      \multicolumn{1}{c}{\textbf{PXrLL}} & \multicolumn{1}{c}{\textbf{InitPX}} \\
        
  \textbf{Problem (size)}     & \textbf{[\%]}     & \textbf{Func / Surr / ratio}             & \textbf{[\%]}                 & \textbf{Total}      & $b^*_g(\vec{x})$    & \textbf{Miss.} & \textbf{Disc.}     & \textbf{Total}     & \textbf{Regular}        & \textbf{Miss.}                  & \textbf{Disc.}             & \textbf{LL}      & \textbf{FIHC}                 & \textbf{Circ}                    & \textbf{Disc} &  \textbf{Disc} \\
\hline
\textbf{nkLand (800)}  & 100     & 1.1E+7 / \textbf{8.0E+7} / \textbf{88}    & 100                   & 1.1E+7   & 7.6E+5 & 1.0E+7                   & 6.1E+4          & 1.6E+5           & 8.1E+4     & 7.8E+4               & 1.9E+3          & 7.9E+2   & 3984                & 43                     & 5 (28)    & 4 (80)    \\
\textbf{ISG (1936)} & 60      & 1.9E+7 / \textbf{4.8E+8} / \textbf{96}    & 100                    & 1.9E+7   & 1.9E+5 & 1.9E+7                   & 7.0E+4          & 3.9E+5           & 2.9E+5     & 9.7E+4               & 3.2E+2          & 8.9E+2   & 3768                & 0                      & 5 (77)    & 11 (95)    \\
\textbf{m3s (75)}   & 90      & 1.4E+5 / 4.7E+4 / 25    & 93                  & 2.1E+5   & 1.3E+5 & 7.6E+4                   & 5.1E+3          & 1.6E+1           & 1.4E+1     & 2.0E+0               & 0.0E+0          & 1.9E+2   & 740                 & 0                      & 0 (-)   & 5 (91)    \\
\textbf{mkDec3 (82)}   & 100     & 3.6E+4 / 3.4E+4 / 49    & 85                   & 3.6E+4   & 1.6E+3 & 3.3E+4                   & 1.3E+3          & 1.2E+2           & 3.8E+1     & 3.6E+1               & 2.1E+1          & 1.4E+2   & 129                 & 0                      & 1 (100)   & 4 (78)    \\
\textbf{Bm10o1 (198)}  & 100     & 1.5E+6 / \textbf{1.2E+7} / \textbf{89}    & 100                   & 8.6E+5   & 4.0E+5 & 4.5E+5                   & 1.2E+4          & 6.6E+5           & 4.1E+5     & 2.4E+5               & 8.8E+1          & 1.3E+3   & 981                 & 1                      & 4 (100)   & 5 (83)    \\
\textbf{nBm10o3 (203)}  & 100     & 2.8E+6 / \textbf{1.7E+7} / \textbf{86}    & 100                   & 2.0E+6   & 1.4E+6 & 6.3E+5                   & 1.4E+4          & 7.7E+5           & 4.3E+5     & 3.3E+5               & 1.9E+3          & 1.7E+3   & 1186                & 14                     & 19 (88)    & 13 (100)   \\
        \end{tabular}
\end{table*}

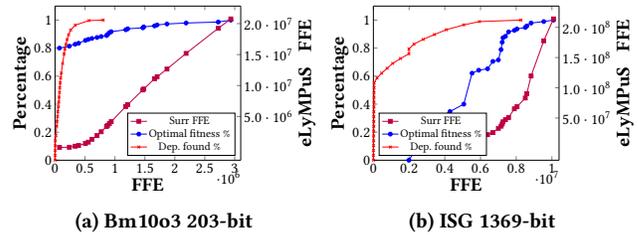
\begin{figure}[ht]
    \begin{subfigure}[b]{0.49\linewidth}
		\resizebox{\linewidth}{!}{%
			\tikzset{every mark/.append style={scale=1}}
			\begin{tikzpicture}
            
			\begin{axis}[
            label style={font=\fontsize{20}{0}\selectfont},
            yticklabel style={font=\fontsize{16}{0}\selectfont},
            xticklabel style={font=\fontsize{16}{0}\selectfont},
            legend style={font=\fontsize{20}{0}\selectfont},
              axis y line*=left,
                ymin=0, ymax=1.1,
              xlabel=\textbf{FFE},
              ylabel=\textbf{Percentage},
			xmin=0,
			xmax=3100000
            ]
   
            \addplot[
			color=blue,
			mark=*
			]
			coordinates {
				%(73917,116)(238278,118)(334873,120)(392904,121)(507195,124)(555151,125)(615759,126)(701009,127)(770524,128)(859850,129)(875933,130)(907165,132)(937928,133)(1180690,135)(1213461,136)(1464821,137)(1481872,138)(1661552,139)(1704063,140)(1868680,141)(2187968,142)(2724551,143)(2934707,145)
                (73917,0.8)(238278,0.813793103448276)(334873,0.827586206896552)(392904,0.83448275862069)(507195,0.855172413793103)(555151,0.862068965517241)(615759,0.868965517241379)(701009,0.875862068965517)(770524,0.882758620689655)(859850,0.889655172413793)(875933,0.896551724137931)(907165,0.910344827586207)(937928,0.917241379310345)(1180690,0.931034482758621)(1213461,0.937931034482759)(1464821,0.944827586206897)(1481872,0.951724137931034)(1661552,0.958620689655172)(1704063,0.96551724137931)(1868680,0.972413793103448)(2187968,0.979310344827586)(2724551,0.986206896551724)(2934707,1)

			};\label{plot_one}

            \addplot[
			color=red,
			mark=x
			] 
  coordinates{
    (419,0.000821018062397373)(1170,0.0344827586206897)(1868,0.0681444991789819)(2865,0.103448275862069)(3632,0.137931034482759)(13718,0.175697865353038)(27615,0.243021346469622)(33499,0.277504105090312)(41231,0.311986863711002)(46883,0.346469622331691)(55332,0.380952380952381)(63025,0.415435139573071)(67402,0.44991789819376)(74529,0.483579638752053)(79213,0.518062397372742)(85713,0.552545155993432)(94083,0.587027914614122)(104072,0.621510673234811)(126559,0.69047619047619)(150845,0.758620689655172)(164649,0.792282430213465)(182631,0.826765188834154)(200565,0.861247947454844)(226788,0.895730706075534)(260873,0.929392446633826)(336725,0.963875205254516)(566192,0.996715927750411)(802970,1)
}; \label{plot_two}
			\end{axis}

            \begin{axis}[
            yticklabel style={font=\fontsize{16}{0}\selectfont},
            legend style={font=\fontsize{12}{0}\selectfont},
              axis y line*=right,
              axis x line=none,
              scaled ticks = false,
              ytick={5000000,10000000,15000000,20000000},
			     yticklabels={$5.0\cdot 10^6$,$1.0\cdot 10^7$,$1.5\cdot 10^7$,$2.0\cdot 10^7$},
              ylabel near ticks, yticklabel pos=right,ylabel style={font=\fontsize{20}{0}\selectfont},
              ylabel=\textbf{eLyMPuS\textcolor{white}{o} FFE},
              legend pos=south east,
              xmin=0,
			     xmax=3100000
            ]

\addplot[
			color=purple,
			mark=square*
			] 
  coordinates{
    (73953,6818)(238278,23380)(334873,194587)(392904,314003)(507195,637533)(555151,849856)(615759,1308710)(701009,1913413)(770524,2530625)(859850,3385876)(875933,3542312)(907165,3854282)(937928,4166320)(1180690,6600742)(1213461,6915367)(1464821,9279866)(1481872,9438012)(1661552,11100448)(1704063,11409558)(1868680,12661031)(2187968,15189480)(2724551,19240837)(2934707,20755873)
}; 

\addlegendimage{/pgfplots/refstyle=plot_one}\addlegendentry{Surr FFE}\addlegendentry{Optimal fitness \%}\addlegendimage{/pgfplots/refstyle=plot_two}\addlegendentry{Dep. found \%}

            \end{axis}
			\end{tikzpicture}
		}
		\caption{Bm10o3 203-bit}
		\label{fig:res:polinomsRuns:bm10o3}
	\end{subfigure}
    \begin{subfigure}[b]{0.49\linewidth}
		\resizebox{\linewidth}{!}{%
			\tikzset{every mark/.append style={scale=1}}
			\begin{tikzpicture}
            
			\begin{axis}[
            label style={font=\fontsize{20}{0}\selectfont},
            yticklabel style={font=\fontsize{16}{0}\selectfont},
            xticklabel style={font=\fontsize{16}{0}\selectfont},
            legend style={font=\fontsize{20}{0}\selectfont},
              axis y line*=left,
                ymin=0, ymax=1.1,
              xlabel=\textbf{FFE},
              ylabel=\textbf{Percentage},
			xmin=0,
			xmax=10400000
            ]
   
            \addplot[
			color=blue,
			mark=*
			]
			coordinates {
				%(1,0)(140,0.0642292490118577)(308,0.123517786561265)(436,0.188735177865613)(616,0.25197628458498)(836,0.308300395256917)(1030,0.357707509881423)(1244,0.41798418972332)(1440,0.481225296442688)(1728,0.538537549407115)(2006,0.591897233201581)(2238,0.648221343873518)(2696,0.700592885375494)(24605,0.744071146245059)(27180,0.786561264822134)(559494,0.8300395256917)(1056107,0.870553359683794)(3017899,0.91699604743083)(5525668,0.964426877470356)(8832413,0.99802371541502)(9542012,0.99901185770751)(10086441,1)
                (1986812,0)(3414912,0.284210526315789)(4288329,0.347368421052632)(5062583,0.4)(5525668,0.621052631578947)(5899123,0.642105263157895)(6397256,0.652631578947368)(6693470,0.705263157894737)(7000981,0.715789473684211)(7151720,0.789473684210526)(7214896,0.842105263157895)(7261750,0.873684210526316)(7397853,0.884210526315789)(7572570,0.91578947368421)(7895779,0.926315789473684)(8046720,0.936842105263158)(8508544,0.947368421052632)(8593019,0.957894736842105)(8833494,0.978947368421053)(9543291,0.989473684210526)(10088593,1)

			};\label{plot_one}

            \addplot[
			color=red,
			mark=x
			] 
  coordinates{
(2757,0.000375375375375375)(4636,0.0349099099099099)(6575,0.0694444444444444)(11616,0.103978978978979)(13624,0.138513513513514)(15635,0.173048048048048)(17688,0.207582582582583)(19722,0.242117117117117)(21751,0.276651651651652)(23822,0.311186186186186)(26565,0.345720720720721)(32290,0.37987987987988)(34435,0.414414414414414)(36592,0.448948948948949)(38804,0.483483483483483)(41074,0.518018018018018)(44113,0.552552552552553)(205057,0.587087087087087)(462558,0.621621621621622)(770810,0.656156156156156)(1081233,0.690690690690691)(1517158,0.725225225225225)(1988576,0.759384384384384)(1991190,0.793918918918919)(2272493,0.828453453453453)(2791198,0.862612612612613)(3417690,0.896771771771772)(4084727,0.931306306306306)(5063901,0.965840840840841)(5941779,0.989489489489489)(8265284,1)

}; \label{plot_two}
			\end{axis}

            \begin{axis}[
            yticklabel style={font=\fontsize{16}{0}\selectfont},
            legend style={font=\fontsize{12}{0}\selectfont},
              axis y line*=right,
              axis x line=none,
              scaled ticks = false,
              ytick={50000000,100000000,150000000,200000000},
			     yticklabels={$5.0\cdot 10^7$,$1.0\cdot 10^8$,$1.5\cdot 10^8$,$2.0\cdot 10^8$},
              ylabel near ticks, yticklabel pos=right,ylabel style={font=\fontsize{20}{0}\selectfont},
              ylabel=\textbf{eLyMPuS\textcolor{white}{o} FFE},
              legend pos=south west,
              xmin=0,
			     xmax=10400000
            ]

\addplot[
			color=purple,
			mark=square*
			] 
  coordinates{
    (1986812,1405182)(3414912,3510054)(4288329,6252791)(5062583,9136822)(5525668,12303314)(5899123,15601442)(6397256,22444884)(6693470,25868338)(7000981,32886715)(7151720,36392572)(7214896,40028503)(7261750,43683599)(7397853,47245665)(7572570,50742039)(7895779,65179418)(8046720,68736065)(8508544,83028369)(8593019,90326500)(8833494,119495776)(9543291,177690595)(10088593,213910228)
}; 

\addlegendimage{/pgfplots/refstyle=plot_one}\addlegendentry{Surr FFE}\addlegendentry{Optimal fitness \%}\addlegendimage{/pgfplots/refstyle=plot_two}\addlegendentry{Dep. found \%}

            \end{axis}
			\end{tikzpicture}
		}
		\caption{ISG 1369-bit}
		\label{fig:res:polinomsRuns:isg}
	\end{subfigure}

	%\hspace{10.11 cm}
	
	\caption{Representative OLyMPuS runs. The increase of the best-found solution, discovered dependency percentage, and the FFE savings brought by eLyMPuS in the experiment run (X axis -- FFE of the true function).}
	\label{fig:res:polinomsRuns}
\end{figure}

\begin{table*}[]
\caption{General comparison. The largest problem size solved with success rate above 80\% (median FFE until finding the best solution reported). Extended results: Tables S-VI and S-VII (supplementary material).} 
    \label{tab:polipmos:general}
\scriptsize
\begin{tabular}{l|rrr|rrr|rrr|rrr|rrr|rrr}
         &    \multicolumn{3}{c}{\textbf{OLyMPuS}} &    \multicolumn{3}{c}{\textbf{P3}}      &   \multicolumn{3}{c}{\textbf{P3-FIHCwLL}} &    \multicolumn{3}{c}{\textbf{P3-eLyMPuS}} &   \multicolumn{3}{c}{\textbf{LT-GOMEA-FIHCwLL}} &   \multicolumn{3}{c}{\textbf{LT-GOMEA-eLyMPuS}} \\
         
         &  \textbf{Size}    & \textbf{Succ}         & \textbf{FFE}            &  \textbf{Size}    & \textbf{Succ}         & \textbf{FFE}&  \textbf{Size}    & \textbf{Succ}         & \textbf{FFE}&  \textbf{Size}    & \textbf{Succ}         & \textbf{FFE}&  \textbf{Size}    & \textbf{Succ}         & \textbf{FFE}&  \textbf{Size}    & \textbf{Succ}         & \textbf{FFE}   \\
\hline
\textbf{nkLand}   & \textbf{\textit{1000}} & 83              & \textbf{\textit{1.8E+7}}         & 400  & 90      & 8.5E+6 & 600  & 93              & 7.2E+6        & \textbf{\textit{1000}} & 83                  & \textbf{\textit{9.0E+6}}            & 400  & 80                & 1.0E+7           & \textbf{\textit{1000}} & 93                    & \textbf{\textit{1.3E+7}}               \\
\textbf{ISG}      & 1369 & 100             & 9.6E+6         & 1936 & 100     & 6.7E+6 & \textbf{1936} & 97              & 7.5E+6        & \textbf{\textit{1936}} & 100                 & \textbf{\textit{2.2E+6}}            & 1024 & 97                & 6.1E+6           & 1369 & 93                    & 5.2E+6               \\
\textbf{m3s}      & 75   & 90              & 1.4E+5         & \textbf{\textit{100}}  & 100     & \textbf{\textit{1.6E+5}} & \textbf{\textit{100}}  & 97              & \textbf{\textit{2.1E+5}}        & \textbf{\textit{100}}  & 97                  & \textbf{\textit{1.8E+5}}            & 75   & 97                & 1.3E+5           & 75   & 93                    & 9.2E+4               \\
\textbf{mkBim10}  & \textbf{\textit{322}}   & 100             & 3.6E+6         & N/A  & N/A     & N/A    & 162   & 80              & 1.2E+7        & 82   & 100                 & 2.5E+6            & 162   & 97                & 1.4E+7           & 162   & 100                   & 9.4E+6               \\
\textbf{mkDec3}   & 102  & 100             & 4.5E+4         & \textbf{\textit{102}}  & 100     & \textbf{\textit{5.8E+3}} & \textbf{\textit{102}}  & 100             & \textbf{\textit{6.3E+3}}        & \textbf{\textit{102}}  & 100                 & \textbf{\textit{5.5E+3}}            & 102  & 100               & 1.2E+4           & 102  & 100                   & 1.0E+4               \\
\textbf{Bim10}    & 400  & 100             & \textbf{\textit{3.2E+6}}         & 50   & 100     & 1.1E+6 & 400  & 90              & 1.3E+7        & 400  & 95                  & 9.4E+6            & 400  & 95                & 9.8E+6           & 400  & 100                   & 1.4E+7               \\
\textbf{Bim10o1}  & \textbf{\textit{297}}  & 100             & \textbf{\textit{2.9E+6}}         & 45   & 100     & 2.4E+6 & 99   & 100             & 3.5E+6        & 99   & 100                 & 3.2E+6            & 198  & 97                & 1.1E+7           & 297  & 80                    & 1.6E+7               \\
\textbf{Bim10o2}  &   \textbf{\textit{400}}   & 95              & 5.5E+6         & 48   & 95      & 6.8E+6 & 104  & 95              & 5.4E+6        & 104  & 100                 & 5.5E+6            & 104  & 100               & 4.7E+6           & 200  & 90                    & 1.3E+7               \\
\textbf{Bim10o3}  & \textbf{\textit{301}}  & 87              & 4.4E+6         & N/A  & N/A     & N/A    & 98   & 93              & 8.3E+6        & 203  & 97                  & 8.2E+6            & 98   & 100               & 9.0E+6           & 98   & 100                   & 4.4E+6               \\
%\textbf{Bim10o4}  & 102  & 100             & \textbf{\textit{2.3E+6}}         & N/A  & N/A     & N/A    & 102  & 83              & 8.8E+6        & 102  & 93                  & 8.4E+6            & 78   & 90                & 8.8E+6           & 102  & 93                    & 1.8E+7               \\
\textbf{nBim10}   & 400  & 100             & \textbf{\textit{2.7E+6}}         & 50   & 100     & 9.9E+5 & 400  & 100             & \textbf{\textit{4.3E+6}}        & 400  & 100                 & \textbf{\textit{3.2E+6}}            & 400  & 100               & 9.7E+6           & 400  & 100                   & 7.0E+6               \\
\textbf{nBim10o1} & 297  & 100             & \textbf{\textit{3.2E+6}}         & 45   & 100     & 2.0E+6 & 198  & 100             & 6.1E+6        & 297  & 80                  & 1.3E+7            & 198  & 100               & 8.3E+6           & 297  & 100                   & 9.7E+6               \\
\textbf{nBim10o2} & \textbf{\textit{400}}  & 90              & 6.1E+6         & 48   & 100     & 4.2E+6 & 200  & 85              & 9.3E+6        & 104  & 100                 & 2.1E+6            & 200  & 100               & 9.3E+6           & 304  & 85                    & 1.2E+7               \\
\textbf{nBim10o3} & \textbf{\textit{301}}  & 83              & 4.7E+6         & 49   & 100     & 4.3E+6 & 98   & 100             & 2.4E+6        & 98   & 100                 & 3.0E+6            & 98   & 100               & 4.5E+6           & 203  & 97                    & 1.5E+7               \\
%\textbf{nBim10o4} & 198  & 100             & \textbf{\textit{3.0E+6}}         & 48   & 100     & 3.6E+6 & 102  & 100             & 2.3E+6        & 102  & 100                 & 3.3E+6            & 102  & 100               & 4.9E+6           & 198  & 87                    & 1.4E+7               \\
\textbf{Dec5}   & 200  & 100             & 5.8E+6         & 200  & 100     & \textbf{\textit{6.8E+4}} & 200  & 100             & \textbf{\textit{9.3E+4}}        & 200  & 100                 & \textbf{\textit{5.8E+4}}            & 200  & 100               & \textbf{\textit{1.6E+5}}           & 200  & 100                   & \textbf{\textit{1.5E+5}}               \\
\textbf{Dec5o1}   & 200  & 100             & 4.4E+6         & 200  & 100     & \textbf{\textit{1.4E+5}} & 200  & 100             & \textbf{\textit{1.6E+5}}        & 200  & 100                 & \textbf{\textit{1.0E+5}}            & 200  & 100               & 2.3E+6           & 200  & 100                   & 7.9E+5               \\
\textbf{Dec5o2}   & 201  & 100             & 4.5E+6         & 201  & 100     & \textbf{\textit{1.6E+5}} & 201  & 100             & \textbf{\textit{2.2E+5}}        & 201  & 100                 & \textbf{\textit{1.5E+5}}            & 201  & 100               & 4.6E+6           & 201  & 100                   & 1.5E+6\\
\hline
\textbf{Most effective} &    \multicolumn{3}{c}{\textbf{10}} &    \multicolumn{3}{c}{\textbf{5}}      &   \multicolumn{3}{c}{\textbf{6}} &    \multicolumn{3}{c}{\textbf{8}} &   \multicolumn{3}{c}{\textbf{1}} &   \multicolumn{3}{c}{\textbf{2}} 
\end{tabular}
\end{table*}

\begin{figure*}[h]
    \begin{subfigure}[b]{0.16\linewidth}
		\resizebox{\linewidth}{!}{%
			\tikzset{every mark/.append style={scale=2.5}}
			\begin{tikzpicture}
			\begin{axis}[%
            legend entries={P3, LT-GOMEA-eLyMPuS, P3-FIHCwLL, LT-GOMEA-eLyMPuS, P3-eLyMPuS, OLyMPuS},
            legend columns=-1,
            legend to name=named,
			%legend columns=-1,
            %legend entries={dgGA,cGOMEA,P3,P3-DLED,LT-GOMEA-DLED},
            %legend to name=named,
            %legend columns=-1,
            %legend entries={LT-GOMEA-DLED,LT-GOMEA-SLL,P3-DLED,P3-SLL},
            %legend to name=named,
			xtick={1,2,3,4,5,6},
			xticklabels={200,300,400,600,800,1000},
			xmin=0.5,
			xmax=6.5,
			%ymode=log,
			%ymin=1e4,
			%ymax=1e9,
			%legend pos=south east,
			%xlabel=\textbf{unitation},
			ylabel=\textbf{FFE until opt.},
			grid,
			grid style=dashed,
			ticklabel style={scale=1.5},
			label style={scale=1.5},
			legend style={font=\fontsize{8}{0}\selectfont}
			]
 \addplot[
			color=brown,
			mark=triangle*,
			]
			coordinates {
				(1,3293183.5)(2,5584697)(3,8458283)
			};

            \addplot[
			color=purple,
			mark=square*,
			]
			coordinates {
				(1,1322284.5)(2,4363997.5)(3,10205367)
			};

            \addplot[
			color=black,
			mark=triangle*,
			]
			coordinates {
				(1,741085.5)(2,1394721)(3,3790814)(4,7160600.5)(5,10976799)
			};
   
            \addplot[
			color=darkgreen,
			mark=square*,
			]
			coordinates {
				(1,1016655)(2,1605788.5)(3,2239329.5)(4,5828520.5)(5,7748752)(6,13286441)
			};

            \addplot[
			color=blue,
			mark=triangle*,
			]
			coordinates {
				(1,781133)(2,911202)(3,1861522)(4,3999085)(5,6635621.5)(6,9029583.5)
			};
   
			\addplot[
			color=red,
			mark=*,
			]
			coordinates {
				(1,716502)(2,1623657.5)(3,3020207)(4,6497554.5)(5,11310511)(6,17873373)
			};

			\end{axis}
			\end{tikzpicture}
		}
		\caption{nkLand}
		\label{fig:scalab:nkLand}
	\end{subfigure}
    \begin{subfigure}[b]{0.16\linewidth}
		\resizebox{\linewidth}{!}{%
			\tikzset{every mark/.append style={scale=2.5}}
			\begin{tikzpicture}
			\begin{axis}[%
            legend entries={P3, LT-GOMEA-eLiPMoS, P3-FIHCwLL, LT-GOMEA-eLiPMoS, P3-eLiPMoS, POLiPMoS},
            legend columns=-1,
            legend to name=named,
			%legend columns=-1,
            %legend entries={dgGA,cGOMEA,P3,P3-DLED,LT-GOMEA-DLED},
            %legend to name=named,
            %legend columns=-1,
            %legend entries={LT-GOMEA-DLED,LT-GOMEA-SLL,P3-DLED,P3-SLL},
            %legend to name=named,
			xtick={1,2,3,4,5},
			xticklabels={484,729,1024,1369,1936},
			xmin=0.5,
			xmax=5.5,
			%ymode=log,
			%ymin=1e4,
			%ymax=1e9,
			%legend pos=south east,
			%xlabel=\textbf{unitation},
			ylabel=\textbf{Fitness},
			grid,
			grid style=dashed,
			ticklabel style={scale=1.5},
			label style={scale=1.5},
			legend style={font=\fontsize{8}{0}\selectfont}
			]

             \addplot[
			color=brown,
			mark=triangle*,
			]
			coordinates {
				(1,415073)(2,862220.5)(3,1806600)(4,3315708.5)(5,6739350)
			};

            \addplot[
			color=purple,
			mark=square*,
			]
			coordinates {
				(1,1254454)(2,4205504)(3,6104265.5)
			};

            \addplot[
			color=black,
			mark=triangle*,
			]
			coordinates {
				(1,387433.5)(2,1117507)(3,2176284.5)(4,4094377.5)(5,7510141.5)
			};
   
            \addplot[
			color=darkgreen,
			mark=square*,
			]
			coordinates {
				(1,504923.5)(2,1152035)(3,2357432)(4,5225910.5)

			};

            \addplot[
			color=blue,
			mark=triangle*,
			]
			coordinates {
				(1,150185.5)(2,291533.5)(3,706875)(4,1071694)(5,2195532.5)
			};
   
			\addplot[
			color=red,
			mark=*,
			]
			coordinates {
				(1,1246645)(2,2872311)(3,5639124.5)(4,9564891)
			};

			\end{axis}
			\end{tikzpicture}
		}
		\caption{\textbf{ISG}}
		\label{fig:scalab:isg}
	\end{subfigure}
    \begin{subfigure}[b]{0.16\linewidth}
		\resizebox{\linewidth}{!}{%
			\tikzset{every mark/.append style={scale=2.5}}
			\begin{tikzpicture}
			\begin{axis}[%
            legend entries={P3, LT-GOMEA-eLiPMoS, P3-FIHCwLL, LT-GOMEA-eLiPMoS, P3-eLiPMoS, POLiPMoS},
            legend columns=-1,
            legend to name=named,
			%legend columns=-1,
            %legend entries={dgGA,cGOMEA,P3,P3-DLED,LT-GOMEA-DLED},
            %legend to name=named,
            %legend columns=-1,
            %legend entries={LT-GOMEA-DLED,LT-GOMEA-SLL,P3-DLED,P3-SLL},
            %legend to name=named,
			xtick={1,2,3},
			xticklabels={82,162,322},
			xmin=0.5,
			xmax=3.5,
			ymode=log,
			%ymin=1e4,
			%ymax=1e9,
			%legend pos=south east,
			%xlabel=\textbf{unitation},
			ylabel=\textbf{FFE until opt.},
			grid,
			grid style=dashed,
			ticklabel style={scale=1.5},
			label style={scale=1.5},
			legend style={font=\fontsize{8}{0}\selectfont}
			]

            \addplot[
			color=brown,
			mark=triangle*,
			]
			coordinates {
				(1,2835666.5)(2,11633974.5)
			};

            \addplot[
			color=purple,
			mark=square*,
			]
			coordinates {
				(1,3314635)(2,13688309.5)
			};

            \addplot[
			color=black,
			mark=triangle*,
			]
			coordinates {
				(1,2835666.5)(2,11633974.5)
			};
   
            \addplot[
			color=darkgreen,
			mark=square*,
			]
			coordinates {
				(1,2221777.5)(2,9350695.5)
			};

            \addplot[
			color=blue,
			mark=triangle*,
			]
			coordinates {
				(1,2529832)

			};
   
			\addplot[
			color=red,
			mark=*,
			]
			coordinates {
				(1,323282.5)(2,1118094)(3,3637915)
			};
            
			\end{axis}
			\end{tikzpicture}
		}
		\caption{\textbf{mkBim10}}
		\label{fig:scalab:mkBim10}
	\end{subfigure} 
%-----------------SECOND ROW-----------------------------
 \begin{subfigure}[b]{0.16\linewidth}
		\resizebox{\linewidth}{!}{%
			\tikzset{every mark/.append style={scale=2.5}}
			\begin{tikzpicture}
			\begin{axis}[%
            legend entries={P3, LT-GOMEA-eLiPMoS, P3-FIHCwLL, LT-GOMEA-eLiPMoS, P3-eLiPMoS, POLiPMoS},
            legend columns=-1,
            legend to name=named,
			%legend columns=-1,
            %legend entries={dgGA,cGOMEA,P3,P3-DLED,LT-GOMEA-DLED},
            %legend to name=named,
            %legend columns=-1,
            %legend entries={LT-GOMEA-DLED,LT-GOMEA-SLL,P3-DLED,P3-SLL},
            %legend to name=named,
			xtick={1,2,3,4,5},
			xticklabels={48,104,200,304,400},
			xmin=0.5,
			xmax=5.5,
			%ymode=log,
			%ymin=1e4,
			%ymax=1e9,
			%legend pos=south east,
			%xlabel=\textbf{unitation},
			ylabel=\textbf{Fitness},
			grid,
			grid style=dashed,
			ticklabel style={scale=1.5},
			label style={scale=1.5},
			legend style={font=\fontsize{8}{0}\selectfont}
			]

              \addplot[
			color=brown,
			mark=triangle*,
			]
			coordinates {
				(1,6786018)
			};

            \addplot[
			color=purple,
			mark=square*,
			]
			coordinates {
				(1,1515118)(2,4719454)
			};

            \addplot[
			color=black,
			mark=triangle*,
			]
			coordinates {
				(1,1006105)(2,5385596.5)
			};
   
            \addplot[
			color=darkgreen,
			mark=square*,
			]
			coordinates {
				(1,1199488.5)(2,5632869.5)(3,13023888.5)

			};

            \addplot[
			color=blue,
			mark=triangle*,
			]
			coordinates {
				(1,829756)(2,5505784.5)
			};
   
			\addplot[
			color=red,
			mark=*,
			]
			coordinates {
				(1,149701)(2,565401.5)(3,1743311.5)(4,3749709)(5,5519853)
			};

			\end{axis}
			\end{tikzpicture}
		}
		\caption{\textbf{Bim10o2}}
		\label{fig:scalab:Bim10o2}
	\end{subfigure}
    \begin{subfigure}[b]{0.16\linewidth}
		\resizebox{\linewidth}{!}{%
			\tikzset{every mark/.append style={scale=2.5}}
			\begin{tikzpicture}
			\begin{axis}[%
            legend entries={P3, LT-GOMEA-eLiPMoS, P3-FIHCwLL, LT-GOMEA-eLiPMoS, P3-eLiPMoS, POLiPMoS},
            legend columns=-1,
            legend to name=named,
			%legend columns=-1,
            %legend entries={dgGA,cGOMEA,P3,P3-DLED,LT-GOMEA-DLED},
            %legend to name=named,
            %legend columns=-1,
            %legend entries={LT-GOMEA-DLED,LT-GOMEA-SLL,P3-DLED,P3-SLL},
            %legend to name=named,
			xtick={1,2,3,4,5,6},
			xticklabels={49,77,98,203,301},
			xmin=0.5,
			xmax=5.5,
			%ymode=log,
			%ymin=1e4,
			%ymax=1e9,
			%legend pos=south east,
			%xlabel=\textbf{unitation},
			ylabel=\textbf{Fitness},
			grid,
			grid style=dashed,
			ticklabel style={scale=1.5},
			label style={scale=1.5},
			legend style={font=\fontsize{8}{0}\selectfont}
			]

           \addplot[
			color=brown,
			mark=triangle*,
			]
			coordinates {
				(1,4342611.5)
			};

            \addplot[
			color=purple,
			mark=square*,
			]
			coordinates {
				(1,1074295.5)(2,3293100)(3,4523830.5)
			};

            \addplot[
			color=black,
			mark=triangle*,
			]
			coordinates {
				(1,489870.5)(2,1292067.5)(3,2364287)
			};
   
            \addplot[
			color=darkgreen,
			mark=square*,
			]
			coordinates {
				(1,854553)(2,2492220)(3,3429033.5)(4,14824776)
			};

            \addplot[
			color=blue,
			mark=triangle*,
			]
			coordinates {
				(1,581247)(2,1367082.5)(3,3016178.5)

			};
   
			\addplot[
			color=red,
			mark=*,
			]
			coordinates {
				(1,209199.5)(2,569440.5)(3,786350.5)(4,2762039.5)(5,4671695.5)

			};
            
			\end{axis}
			\end{tikzpicture}
		}
		\caption{\textbf{nBim10o3}}
		\label{fig:scalab:nBim10o3}
	\end{subfigure}
    \begin{subfigure}[b]{0.16\linewidth}
		\resizebox{\linewidth}{!}{%
			\tikzset{every mark/.append style={scale=2.5}}
			\begin{tikzpicture}
			\begin{axis}[%
            legend entries={P3, LT-GOMEA-FIHCwLL, P3-FIHCwLL, LT-GOMEA-eLyMPuS, P3-eLyMPuS, OLyMPuS},
            legend columns=-1,
            legend to name=named,
			%legend columns=-1,
            %legend entries={dgGA,cGOMEA,P3,P3-DLED,LT-GOMEA-DLED},
            %legend to name=named,
            %legend columns=-1,
            %legend entries={LT-GOMEA-DLED,LT-GOMEA-SLL,P3-DLED,P3-SLL},
            %legend to name=named,
			xtick={1,2,3,4,5},
			xticklabels={28,48,72,100,200},
			xmin=0.5,
			xmax=5.5,
			%ymode=log,
			%ymin=1e4,
			%ymax=1e9,
			%legend pos=south east,
			%xlabel=\textbf{unitation},
			ylabel=\textbf{Fitness},
			grid,
			grid style=dashed,
			ticklabel style={scale=1.5},
			label style={scale=1.5},
			legend style={font=\fontsize{8}{0}\selectfont}
			]

             \addplot[
			color=brown,
			mark=triangle*,
			]
			coordinates {
				(1,12002.5)(2,20833)(3,28400.5)(4,53693.5)(5,141628)
			};

            \addplot[
			color=purple,
			mark=square*,
			]
			coordinates {
				(1,71667.5)(2,253061)(3,614504)(4,1104538)(5,2289544)
			};

            \addplot[
			color=black,
			mark=triangle*,
			]
			coordinates {
				(1,13659.5)(2,23687)(3,40219)(4,56493.5)(5,158441)
			};
   
            \addplot[
			color=darkgreen,
			mark=square*,
			]
			coordinates {
				(1,18801.5)(2,69235.5)(3,157263)(4,311947)(5,787904.5)
			};

            \addplot[
			color=blue,
			mark=triangle*,
			]
			coordinates {
				(1,5978)(2,16735.5)(3,25347)(4,38026)(5,99515)
			};
   
			\addplot[
			color=red,
			mark=*,
			]
			coordinates {
				(1,136592.5)(2,294651.5)(3,499429.5)(4,917909)(5,4386716)
			};
            
			\end{axis}
			\end{tikzpicture}
		}
		\caption{\textbf{Dec5o1}}
		\label{fig:scalab:Dec5o1}
	\end{subfigure}
	
    \hspace{5 cm}
	\pgfplotslegendfromname{named}
	%\ref{named2}

	\caption{Scalability for chosen problems}
	\label{fig:scalab}
\end{figure*}
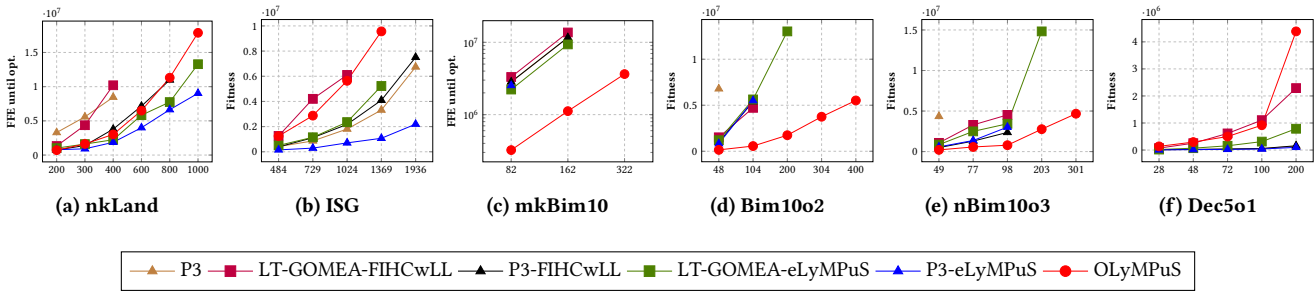

Fig. \ref{fig:res:polinomsRuns} presents two representative OLyMPuS runs. The X-axis presents the consumption of the FFE-based computation budget. In Fig. \ref{fig:res:polinomsRuns:bm10o3}, all dependencies are quickly found (the red curve reaches 1). When this happens, the number of eLyMPuS-saved FFEs increases quickly (brown curve) because no missing linkages are found, and the costs of checking them drop thanks to the $1/v^{th}$ strategy. Nevertheless, the FFE budget required to find the optimal solution was relatively large, indicating that finding $\VIG$ and constructing a surrogate make the optimization easier but not necessarily easy.\par

Fig. \ref{fig:res:polinomsRuns:isg} presents a run in which high-quality solutions were found before finding all dependencies. Also, the eLyMPuS-saved FFE number starts increasing before the red curve ends. After finding $\VIG$, the optimal solution is found very quickly.\par

Table \ref{tab:polipmos:detailed} presents detailed statistics of the  FIHC-based costs of OLyMPuS for the chosen instance groups. For many problems, the eLyMPuS-based savings are significant (over 90\% of evaluations can be handled by eLyMPuS). eLyMPuS discovers a vast majority of dependencies, and the highest cost is the missing linkage detection, i.e., verifying whether the partial comparison output returned by eLyMPuS is correct. The cost of linkage discovery is lower by one or even three orders of magnitude. Similar observations can be made for PXrLL, but its costs seem negligible when compared to FIHC-eLyMPuS. The comparison of other OLyMPuS versions is presented in Table S-V (supplementary material).\par

Table \ref{tab:polipmos:detailed} shows that the percentage of dependencies found by PXrLL is negligible compared to eLyMPuS, leading to the conclusion that missing linkage detection and linkage discovery can be removed from PXrLL. Surprisingly, it is false. Table S-V (supplementary material) presents a detailed comparison of the standard OLyMPuS version and the version that does not perform any missing-linkage detection (and therefore does not perform any linkage discovery) in PXrLL. We consider only those instance groups for which the median percentage of the found dependencies is below 100\%. In 25 out of 28 instance groups, standard OLyMPuS obtained a higher median of discovered dependencies. This difference was caused by PXrLL. For nkLand and ISG, this tiny difference caused a significant difference in the percentage of successful runs. Thus, even if the number of dependencies discovered by PXrLL and FIHC-eLyMPuS differ by several levels of magnitude, PXrLL seems to discover those dependencies that elude FIHC-eLyMPuS.\par

In Table \ref{tab:polipmos:general}, we show the largest considered instance solved in at least 80\% of the runs by a given optimizer. If more than one optimizer reached the same instance size, then we compare the statistical significance of FFE differences. If the differences are meaningful, then the faster optimizer is considered more effective. Otherwise, more than one optimizer can be found to be the best for a given problem. OLyMPuS was among the most effective optimizers for 10 out of 16 problems and outperforms the other optimizers. Note that the second-most-effective optimizer was P3-eLyMPuS, and P3-FIHCwLL took third place. These results confirm that eLyMPuS offers a range of functionalities that can improve the performance of various optimizers (LT-GOMEA-eLyMPuS outperformed its FIHCwLL and base versions). Finally, these observations are confirmed by the scalability analysis presented in Fig. \ref{fig:scalab}.

\section{Conclusions}
\label{sec:conc}
Here, we propose eLyMPuS, which is based on the non-monotonicity check and applies to problems that can not be modeled by perfect linear surrogates. It enables partial comparisons, i.e., comparing two solutions that differ by one variable. If eLyMPuS is trained using $\eVIG$ such that $\eVIG=\VIG$, it is guaranteed to yield a correct answer. The model-building process is not a separate step but is performed during use, and FFE-based costs are paid only when necessary, making the eLyMPuS highly flexible. Given that it is parameter-less, it may be well-suited to solving real-world problems.\par

eLyMPuS functionality extends beyond a typical surrogate. Indeed, within its routine, it enables low-cost missing-linkage detection and a low-cost linkage-discovery procedure that is guaranteed to discover one missing dependency. It can also tell symmetrical and non-symmetrical dependencies. Finally, the proposed mechanisms are not limited to the binary search space and apply to integer-based solution encoding. They can also be adjusted to other search spaces, which is a future work step. However, the most important future objective is to propose a perfect monotonic surrogate without limitations, i.e., one that can compare any pair of solutions.

\begin{acks}
We would like to thank Peter Bosman for his valuable remarks.
This work was supported by: Polish National Science Centre (NCN), 2022/45/B/ST6/04150 (M.M. Komarnicki, M.W. Przewozniczek);\\ MCIN/AEI/10.13039/501100011033, PID2024-158752OB-I00, the University of Malaga, PAR 4/2023 (F. Chicano); FAPESP, \#2024/15430-5 and \#2024/08485-8, CNPq, \#304640/2024-7 (R. Tin\'os).
\end{acks}

	\bibliographystyle{ACM-Reference-Format}
	\bibliography{LimitedNonMonoSurr} 
	
\end{document}